\documentclass{article}

\usepackage{microtype}
\usepackage{graphicx}
\usepackage{caption}
\usepackage{subcaption}
\usepackage{booktabs}
\usepackage{amsmath}
\usepackage{amssymb}
\usepackage{hyperref}
\usepackage{cleveref}
\usepackage{xcolor}

\usepackage{paperstyle}

\newcommand{\sysname}{MixFold}
\newcommand{\Requant}{\operatorname{Requant}}
\newcommand{\npunameshort}{Qualcomm Hexagon NPU of a Snapdragon~8 Elite Gen~5}
\newcommand{\npuname}{\npunameshort{} (SM8850, Hexagon v81)}
\newcommand{\naivepath}{expand path}
\newcommand{\proposepath}{fold path}

\newcommand{\para}[1]{\noindent \textbf{#1.}}
\usepackage{placeins} %

\AddToHook{begindocument/before}{%
  \papersetsymbol{intern}{*}%
  \AddToHook{env/paperauthorlist/end}{%
    \par\vspace{6pt}%
    {\normalfont\small
      \textsuperscript{1}\,Qualcomm AI Research, Qualcomm Technologies, Inc.,
      San Diego, CA, USA\par\vspace{2pt}%
      \textsuperscript{2}\,Seoul National University, Seoul, Republic of Korea
      \par
    }%
  }%
  \renewcommand{\printAffiliationsAndNotice}[1]{%
    {\let\thefootnote\relax
      \expandafter\long\expandafter\def\csname @makefntext\endcsname##1{%
        \noindent ##1}%
      \footnotetext{%
        \raggedright
        #1\textsuperscript{*}\,Work performed while Seong Hoon Seo was
        an intern at Qualcomm Technologies, Inc.%
        \par\vspace{3pt}%
        Correspondence:
        \href{mailto:vinnkim@qti.qualcomm.com}{\nolinkurl{vinnkim@qti.qualcomm.com}}%
        \par\vspace{5pt}%
        \csname Notice@String\endcsname
      }%
    }%
  }%
  \hypersetup{%
    pdfsubject={Preprint},
    pdfauthor={Vinnam Kim, Seong Hoon Seo, Junhan Kim,
               Jaeseong You, An Chen}
  }%
}

\papertitlerunning{\sysname{}: Sub-4-Bit Mixed-Precision LLM Inference via Output-Channel Grouping on Mobile NPUs}

\begin{document}

\twocolumn[
\papertitle{\sysname{}: Sub-4-Bit Mixed-Precision LLM Inference
      via Output-Channel Grouping on Mobile NPUs}

\begin{paperauthorlist}
\paperauthor{Vinnam Kim}{qti}
\paperauthor{Seong Hoon Seo}{snu,qti,intern}
\paperauthor{Junhan Kim}{qti}
\paperauthor{Jaeseong You}{qti}
\paperauthor{An Chen}{qti}
\end{paperauthorlist}

\paperaffiliation{qti}{ Qualcomm AI Research is an initiative of Qualcomm Technologies, Inc., San Diego, CA, USA }
\paperaffiliation{snu}{ Seoul National University, Seoul, Republic of Korea }

\papercorrespondingauthor{Vinnam Kim}{vinnkim@qti.qualcomm.com}

\paperkeywords{Large Language Models, Quantization, Mobile NPUs, Mixed Precision}

\vskip 0.3in

\begin{abstract}

Sub-4-bit quantization allows mobile devices to run larger LLMs, but its memory savings do not always translate into faster decoding on mobile NPUs. Their matrix units support only a few native integer datatypes, and converting unsupported weight formats on the vector unit can erase the gains from fewer bits. Mixed precision faces an additional constraint: each matrix-unit tile uses one datatype.
\sysname{} executes 3-bit and perfectly symmetric 2-bit weights using native integer operations without expanding the weights to wider datatypes during decoding.
\sysname{} then assigns datatypes to output-channel groups aligned with the matrix-unit tiles, enabling finer-than-layer assignment without disrupting matrix-unit execution.
We compare \sysname{} with llama.cpp Hexagon backend on the \npunameshort{}.
Across the Qwen3 model family, \sysname{} achieves comparable or higher accuracy than llama.cpp's sub-4-bit formats at matched bits per weight, while decoding 1.16--1.41$\times$ faster and generating 1.5--4.3$\times$ more tokens per joule.

\end{abstract}

]

\printAffiliationsAndNotice{}

\section{Introduction}
\label{sec:introduction}

Sub-4-bit quantization, below 4 bits per weight (BPW) on average, allows a mobile device to hold a larger LLM.
Larger LLMs can better support \textit{on-device} AI use cases involving personal data such as contacts, photos, calendar events, and messages.
Two levers can improve accuracy under a sub-4-bit budget.
The first is a better datatype, such as the perfectly symmetric 2-bit grid~\citep{liu2026paretoq} or 3-bit weights.
The second is mixed precision (MP), which assigns more bits to sensitive weights and fewer bits to the rest.
With these additional format choices, MP can further improve accuracy at a given BPW.

However, on mobile NPUs, the most accurate sub-4-bit configurations save memory, but their decode time may not be as fast as their bytes promise.
This is because the matrix unit of a mobile NPU admits only a few datatypes directly in hardware.
We call these datatypes \emph{native}.
In contrast, \emph{non-native} datatypes require converting the weights on the vector unit before the matrix unit execution.
For example, a straightforward approach expands perfectly symmetric 2-bit and 3-bit weights to native 4-bit values on the vector unit before matrix-unit execution.
A GPU could hide such conversion with a powerful general computing capability~\citep{frantar2025marlin,xia2024fp6llm,zeng2025abq}, but on a mobile NPU, one vector-unit thread delivers two to three orders of magnitude less throughput than the matrix unit~\citep{hao2026scaling}.
MP faces a constraint driven by the matrix unit as well.
Each matrix unit tile uses one datatype, so mixing bitwidths within a tile breaks the native layout.

The two levers thus set two requirements.
Every datatype in the mix must execute natively, and each weight tile must hold one bitwidth.
Existing approaches do not fulfill both requirements at once.
llama.cpp converts the weights on the vector unit before every multiplication~\citep{llamacpp}.
T-MAN skips the conversion with lookup tables (LUTs) on the vector unit, but it leaves the matrix unit idle at decode~\citep{wei2025tman}.
\citet{hao2026scaling} align 4-bit quantization groups with the matrix-unit tile, but they still dequantize the weights to floating point.
MP methods assign bitwidths per layer~\citep{dong2020hawq,yao2026gamma}, per channel~\citep{zheng2026mixllm}, or per salient group~\citep{huang2024slim}.
Per-layer assignment keeps each tile in one bitwidth but ignores the sensitivity variation within a layer.
The finer two may assign different bitwidths within a tile.

In this paper, we propose \sysname{}, which satisfies both requirements at the finest granularity permitted by the NPU's tiled architecture.
At decode, \sysname{} executes perfectly symmetric 2-bit weights using native 2-bit codes by folding the correction into the bias.
For 3-bit weights, it uses bit-plane decomposition, processing the 2-bit and 1-bit planes on the matrix and vector units in parallel.
For MP, \sysname{} assigns bitwidths to output-channel groups aligned with these tiles, enabling finer-than-layer assignment aligned with the matrix unit's tile-level scheduling.
This MP strategy also incorporates the non-native datatypes to further enhance the model accuracy at lower BPW.
We test this synergy on \npunameshort{} across the Qwen3 model family.
At matched BPW, \sysname{} achieves comparable or higher accuracy than llama.cpp's sub-4-bit formats, decodes 1.16--1.41$\times$ faster, and generates 1.5--4.3$\times$ more tokens per joule.

This paper makes the following contributions:
\begin{itemize}
    \item Decode paths that run perfectly symmetric 2-bit and 3-bit weights on native integer MMA without converting the weights (\Cref{sec:kernels}).
    \item Mixed precision over hardware-aligned output-channel groups with both non-native formats effective on the lower BPW (\Cref{sec:group-mp}).
    \item An evaluation on a commercial mobile NPU showing that sub-4-bit weights now decode as fast as their bytes promise, that mixes add negligible overhead, and that the richer datatypes and finer granularity raise accuracy at the same BPW (\Cref{sec:experiments}).
\end{itemize}
\section{Background and Motivation}
\label{sec:background}

\begin{figure}[t]
  \centering
  \includegraphics[width=\linewidth]{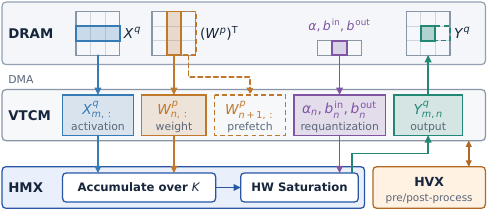}
  \caption{Quantized MMA dataflow on the Hexagon NPU for one $T\times T$
  output tile $Y^q_{m,n}$. DMA stages the activation row block $X^q_{m,:}$,
  the weight tile $W^p_{n,:}$, and the requantization parameters from DRAM in
  VTCM while the next weight tile is prefetched. HVX pre- and
  post-processes VTCM buffers. HMX accumulates over $K$, then applies
  $\alpha_n$, $b^{\mathrm{in}}_n$, and $b^{\mathrm{out}}_n$ with hardware
  saturation.}
  \label{fig:bg-hmx-architecture}
\end{figure}

\begin{figure}[t]
  \centering
  \includegraphics[width=\linewidth]{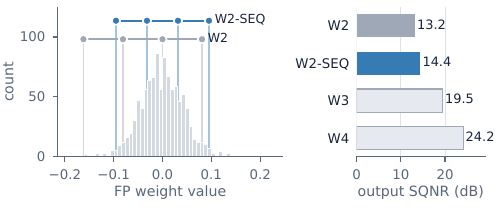}
  \caption{(Left) Histogram of row 0 of the layer-0 gate projection of Qwen3-0.6B, with its W2 and W2-SEQ grids (markers). (Right) Output SQNR of the whole layer on WikiText-2 test activations, with per-channel scales fitted on WikiText-2 train activations.}
  \label{fig:bg-qscheme-sqnr}
\end{figure}

\subsection{Integer MMA Execution on Mobile NPUs}
\label{sec:bg-quantization}

\para{Quantization for Integer MMA}
Weight quantization encodes each weight as a low-bit integer rather than a floating-point (FP) value, which reduces both the memory footprint and the amount of data read per generated token. Quantization also saves energy when the arithmetic operates on the low-bit integers directly. Low-precision integer arithmetic can reduce energy per operation relative to FP arithmetic~\citep{horowitz20141,noh2025jack}, motivating integer matrix multiply-accumulate (MMA) support in mobile NPUs~\citep{park2022multi}.
On \npunameshort{}, \Cref{tab:bg-matmul-energy} shows that a W4A16 integer MMA uses 34--62\% less SoC energy than llama.cpp's Q4\_0~\citep{llamacpp}, which stores the weights at a similar 4.5 bits but performs the matrix multiplication in FP16.

\begin{table}[t]
\centering
\caption{Energy (millijoules, mJ) of one $K=N=2048$ MMA. W4A16 takes 16-bit activation codes in either row-major format (converted inside the kernel) or the tiled HMX layout.}
\label{tab:bg-matmul-energy}
\resizebox{\columnwidth}{!}{%
\begin{tabular}{llrrrr}
\toprule
 & & \multicolumn{2}{c}{$M=32$} & \multicolumn{2}{c}{$M=512$} \\
\cmidrule(lr){3-4}\cmidrule(lr){5-6}
Kernel & MMA & mJ & vs.\ FP16 & mJ & vs.\ FP16 \\
\midrule
llama.cpp Q4\_0 & FP16 & 0.378 & -- & 2.248 & -- \\
W4A16, row-major & INT & 0.156 & $-$59\% & 1.479 & $-$34\% \\
W4A16, tiled & INT & 0.144 & $-$62\% & 1.303 & $-$42\% \\
\bottomrule
\end{tabular}

}%
\end{table}

Low-precision integer arithmetic is realized with uniform weight--activation quantization~\citep{nagel2021white}, which represents both the weights and the activations as integers on a uniform grid. Consider a linear layer $Y=XW^{\mathsf T}+b$, where $Y\in\mathbb{R}^{M\times N}$ is the output corresponding to the input activation matrix $X\in\mathbb{R}^{M\times K}$, $W\in\mathbb{R}^{N\times K}$ is the weight matrix, and $b\in\mathbb{R}^{N}$ is the bias. Here, $M$, $K$, and $N$ denote the number of activation vectors, input channels, and output channels, respectively. Assuming per-tensor asymmetric quantization for the activations and per-channel symmetric quantization for the weights, the activations and the weights are quantized as
\begin{equation}
  \begin{aligned}
    X_{m,k} \approx s_a\bigl(X^q_{m,k}-z_a\bigr),\quad
    W_{n,k} \approx s_{w,n}W^p_{n,k},
  \end{aligned}
  \label{eq:bg-quantization}
\end{equation}
where $X^q_{m,k}$ and $W^p_{n,k}$ are the unsigned $q$-bit and signed $p$-bit integer codes of the activations and the weights, respectively. 
Here, all activations share a scale $s_{a}$ and a zero point $z_{a}$, while all weights in output channel $n$ share a scale $s_{w, n}$.

Using these quantized codes, the integer MMA first computes the raw integer accumulation
\begin{equation}
  P_{m,n}=\sum_{k=1}^{K}X^q_{m,k}W^p_{n,k},
  \label{eq:bg-raw-accumulation}
\end{equation}
which is then requantized to the output code through per-channel rescaling and affine correction:
\begin{equation}
  \begin{aligned}
  Y^q_{m,n}
  &=\Requant\bigl(P_{m,n};\alpha_n,b^{\mathrm{in}}_n\bigr)\\
  &=\operatorname{clip}\bigl(\bigl\lfloor
      \alpha_n(P_{m,n}-b^{\mathrm{in}}_n)+b^{\mathrm{out}}_n
    \bigr\rceil\bigr),
  \end{aligned}
  \label{eq:bg-integer-mma}
\end{equation}
where $\operatorname{clip}(\cdot)$ saturates the result to the output code range $[0,2^q-1]$ and
\begin{equation}
  \begin{aligned}
    \alpha_n = \frac{s_a s_{w,n}}{s_y}, \quad
    b^{\mathrm{in}}_n = z_a\sum_{k=1}^{K}W^p_{n,k}, \quad
    b^{\mathrm{out}}_n = \frac{b_n}{s_y}+z_y.
  \end{aligned}
  \label{eq:bg-conversion-parameters}
\end{equation}
Here, $s_y$ and $z_y$ are the scale and zero point of the output, which is quantized in the same way as the activations. We list only the scale and input bias as arguments of $\Requant$, since \sysname{} mainly modifies these two (\Cref{sec:kernels}); the output bias remains $b^{\mathrm{out}}_n$ unless stated otherwise.

\para{Execution on Mobile NPUs}
\Cref{fig:bg-hmx-architecture} illustrates how a mobile NPU executes Integer MMA, using Qualcomm's Hexagon NPU~\citep{mahurin2023qualcomm} as a representative example. The NPU comprises four components: Hexagon Matrix eXtensions (HMX), a matrix unit that performs the MMA; Hexagon Vector eXtensions (HVX), a vector unit for general-purpose computation; Vector Tightly Coupled Memory (VTCM), a software-managed on-chip memory shared by HMX and HVX; and a direct memory access (DMA) engine that moves data between main memory (DRAM) and VTCM~\citep{qualcomm_hmx_v81,qualcomm_hvx_v79}.

HMX operates on \emph{tiles}: it multiplies an activation tile by a weight tile and accumulates the products into a $T \times T$ output tile.
The weights, activations, and requantization parameters reside in main memory. Since the capacity of the on-chip memory is limited, they are staged by DMA for one output tile at a time, which spans $T$ activation vectors and $T$ output channels, and the data for the next output tile are prefetched during the current MMA~\citep{llamacpp}. For each output tile, the matrix unit loads successive activation and weight tiles along the $K$ dimension into its registers, accumulates their products (see~\eqref{eq:bg-raw-accumulation}), and then applies the requantization with one set of parameters per output channel (see~\eqref{eq:bg-integer-mma} and~\eqref{eq:bg-conversion-parameters}).

We call a format \emph{native} when its datatype and tiled layout match the operands consumed by the matrix unit, thereby leading to a seamless flow of MMA and requantization; otherwise, it is called \emph{non-native}. A prime example of a non-native format is llama.cpp's Q4\_0, which packs 32 consecutive 4-bit integer codes with a scale. This format requires the vector unit to convert weights into a native datatype (such as FP16) before they reach the matrix unit.

\subsection{Challenges in Sub-4-Bit Quantization}
\label{sec:bg-challenges}

\begin{table}[t]
\centering
\caption{Effective weight bandwidth (weight bytes over latency) at $M=1$, aggregated over four projections of Qwen3-4B (QKV+O+Up/Gate+Down), against the DMA ceiling, a bulk copy of the same bytes with no computation at all.}
\label{tab:bg-bandwidth}
\small
\begin{tabular}{lrr}
\toprule
 & GB/s & \% of ceiling \\
\midrule
DMA ceiling & 84.1 & 100 \\
\midrule
Native W4 & 76.4 & 91 \\
Native W4 excluding HMX & 79.2 & 94 \\
\midrule
llama.cpp Q4\_0 & 67.2 & 80 \\
T-MAN INT4 & 71.0 & 84 \\
\bottomrule
\end{tabular}

\end{table}

\para{Non-Native Weight Representations}
Weight bitwidths below 4 bits, namely 3 and 2 bits, reduce the memory footprint further. However, 3-bit weights are non-native, since HMX provides no native MMA for 3-bit weights. Indeed, 3-bit quantization is commonly unsupported across most NPUs, not just Hexagon, because compute units are fundamentally optimized for power-of-two data widths~\citep{xia2024fp6llm,zeng2025abq,park2022multi}. In contrast, 2-bit weights are native, but their grid limits accuracy. The native 2-bit grid is the conventional signed integer grid $\{-2,-1,0,1\}$, which is asymmetric about zero and covers a narrower range of positive weights than of negative ones. Stretched Elastic Quantization (SEQ) was proposed by ParetoQ~\citep{liu2026paretoq} for such extreme low-bit quantization and instead maintains a perfectly symmetric grid $\{-3,-1,1,3\}$ at 2 bits.
We call this format W2-SEQ; unsuffixed W2, W3, and W4 denote the signed integer grids.
ParetoQ reported that W2-SEQ outperforms W2 in per-channel QAT.
\Cref{fig:bg-qscheme-sqnr} also shows that W2-SEQ yields a higher output signal-to-quantization-noise ratio (SQNR) than W2, and W3 comes close to W4.

Even though W3 and W2-SEQ can improve model quality under sub-4-bit quantization, their non-native representations require runtime conversion. Since the W3 grid $\{-4,\ldots,3\}$ and the W2-SEQ grid are both contained in the W4 grid $\{-8,\ldots,7\}$, the vector unit can expand their weights to native W4. 
However, the matrix unit cannot start the MMA until the less powerful vector unit~\citep{hao2026scaling} completes this expansion.
The expanded weights cannot be retained in the limited on-chip memory, while storing W4 copies in DRAM would forfeit the savings of fewer bits.
Consequently, the expansion must be repeated for every forward pass, and its cost is amortized only over $M$ activation vectors: many during prefill, but one or a few during decode.
HVX also serves camera and vision workloads~\citep{lucian2015dspimaging}. With few HVX threads, this expansion falls behind the weight stream at decode.

In \Cref{tab:bg-bandwidth}, the native W4 kernel streams its weights at 91\% of the DMA ceiling.
When the HMX computation is removed, the native W4 kernel reaches 94\% of the DMA ceiling, indicating that HMX computation costs only 3\%.
The work around the HMX, such as the per-call setup and the chunked DMA pipeline, costs the remaining 6\%.
In contrast, llama.cpp's Q4\_0 reaches only 80\% of the DMA ceiling.
At $M=1$, the kernel runs on HVX, which unpacks the 4-bit codes and rescales each block of 32 weights during execution.
T-MAN~\citep{wei2025tman} instead performs decode using lookup tables (LUTs) on HVX.
We run its public 4-bit decode kernel on the DMA pipeline and thread pool of llama.cpp's Hexagon backend.
The resulting kernel reaches 84\% of the DMA ceiling.
Neither approach matches the performance of the native W4 kernel.

\para{Hardware-aligned MP}
MP benefits mobile NPU inference only when latency scales with bits per weight (BPW).
For example, a 3.0-BPW model that mixes W2 and W4 should decode approximately halfway between the uniform W2 and W4 models.
This requires that each datatype MP utilizes achieve byte-proportional execution and that dispatching different formats incur negligible overhead.
Within these constraints, the assignment granularity should be as fine as possible. Layer-wise assignment~\citep{dong2020hawq,yao2026gamma} meets both requirements but ignores sensitivity variation within a layer.
Per-channel assignment~\citep{zheng2026mixllm} captures this variation but may assign different bitwidths to channels within the same weight tile, which the matrix unit cannot process directly.
Regrouping channels by bitwidth may keep each tile in a single bitwidth, but HMX would produce output channels in the regrouped order, requiring an additional pass to restore the original channel order.
Consequently, per-channel assignment introduces repeated conversion or reordering overhead, preventing latency from scaling with BPW.
\par

\section{Executing Non-Native Weights with Native Integer MMA}
\label{sec:kernels}

\begin{figure*}[t]
  \centering
  \includegraphics[width=\textwidth]{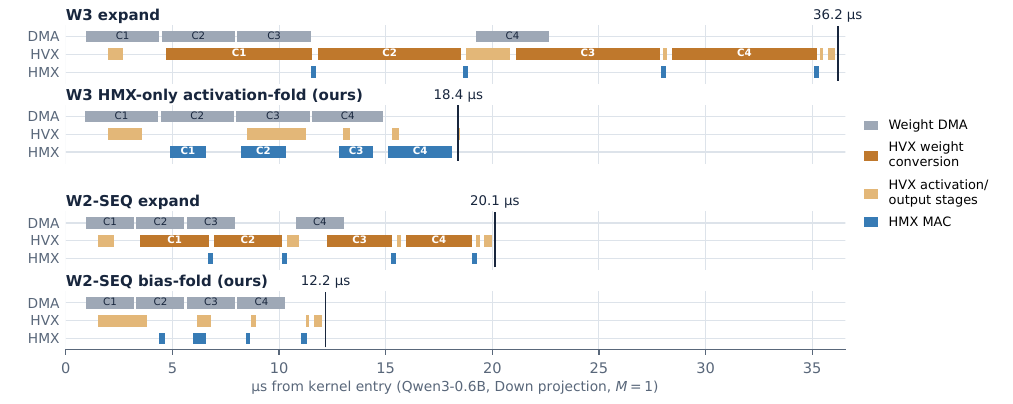}
  \caption{The \proposepath{}s take the HVX workload off the critical path. All kernels utilize single-thread HVX. C1--C4 number the chunks.}
  \label{fig:lowbit-schedule}
\end{figure*}

\begin{table}[t]
\centering
\small\setlength{\tabcolsep}{2pt}\renewcommand{\arraystretch}{1.15}
\caption{Per-call work of each path at its production $M$.
Only the \naivepath{} touches entire $NK$ weights on HVX.}
\label{tab:work-placement}
\begin{tabular*}{\columnwidth}{@{\extracolsep{\fill}}llll@{}}
\toprule
Path & $M$ & HVX work & HMX MACs \\
\midrule
Expand & any & $O(NK)$ & $MNK$ \\
\addlinespace
\multicolumn{4}{@{}l}{W2-SEQ} \\
\quad bias-fold & $1$ & $O(K{+}N)$ & $MNK$ \\
\quad accumulator-fold & $2$--$32$ & $O(M(K{+}S){+}N)$ & $MN(K{+}S)$ \\
\addlinespace
\multicolumn{4}{@{}l}{W3 activation-fold} \\
\quad HVX-assist & $1$ & $O(K)$, $NK/4$ lookups & $MN(K{+}S)$ \\
\quad HMX-only & $2$--$32$ & $O(MK)$ & $2MNK$ \\
\bottomrule
\end{tabular*}
\end{table}
\par

As mentioned in \Cref{sec:bg-challenges}, W2-SEQ and W3 lack a native HMX MMA path.
A naive approach to execute both formats is to expand their weights to signed W4 when loading them into VTCM (\Cref{app:upcast}); we call this the \emph{\naivepath{}}.
However, the \naivepath{} makes every MMA wait for weight conversion, and decode offers little weight reuse to amortize this overhead.
Instead, our approach keeps the weights in native low-bit layouts and folds the remaining work into operands whose cost scales with the activations rather than the weights.
We call these the \emph{\proposepath{}s}.

\subsection{W2-SEQ via a Row-Sum Fold}
\label{sec:symmetric-w2}

\paragraph{Offline encoding.}
W2-SEQ can be encoded in native signed W2 as
$\widetilde W^2=2W^2+\mathbf{1}$. 
We encode and pack the weights as $W^2=(\widetilde W^2-\mathbf{1})/2$ in the native tile layout,
retaining the original SEQ scales and the SEQ input bias
$\widetilde b^{\mathrm{in}}_n=z_a\sum_k\widetilde W^2_{n,k}$.

\paragraph{Runtime correction.}
Using $\widetilde{W}^{2}=2W^{2}+\mathbf{1}$, the W2-SEQ product can be rewritten in terms of the native W2 accumulator $P^{2}_{m,n}$ as 
\begin{equation}
  \sum_{k}X^q_{m,k}\widetilde W^2_{n,k}=2P^{2}_{m,n}+R_m,\qquad
  R_m=\sum_{k}X^q_{m,k}.
  \label{eq:seq-rewrite}
\end{equation}
Here, $R_m$ is the sum of activation row $m$ and thus depends on the input row rather than the output channel.
By folding the factor of two into the requantization scale, we obtain
\begin{equation}
  Y^q_{m,n}\approx\Requant\bigl(P^{2}_{m,n};2\alpha_n,(\widetilde b^{\mathrm{in}}_n - R_m)/2\bigr).
  \label{eq:seq-converted-output}
\end{equation}
This formulation enables HMX to execute W2-SEQ using its native W2 MMA.
However, the required input-bias correction $(\widetilde b^{\mathrm{in}}_n-R_m)/2$ varies across rows, whereas HMX applies a single input bias per output channel to every row in an output tile (see~\eqref{eq:bg-integer-mma}).
A general solution is the \emph{output-correction form} described in \Cref{app:output-correction-form}, which applies the row-dependent correction after $\Requant$ and then saturates the output again.
This approach requires an additional HVX pass over all $M \times N$ outputs and may lose precision because of repeated output saturation.
The two fold paths below avoid both issues by correcting before $\Requant$.

\paragraph{Bias-fold.}
When $M=1$, there is only one activation row, so $R_1$ is shared across all output channels and $(\widetilde b^{\mathrm{in}}_n-R_1)/2$ remains a valid per-channel input bias.
We therefore fold $R_{1}$ into the input bias, which we call the \emph{bias-fold}:
\begin{equation}
    Y^q_{1,n}\approx\Requant\bigl(P^{2}_{1,n};\,2\alpha_n,\,
    (\widetilde b^{\mathrm{in}}_n-R_1)/2\bigr).
  \label{eq:seq-bias-fold}
\end{equation}
Per call, HVX computes $R_{1}$ by reducing the activation row and updates the $N$ per-channel input-bias entries.
HMX then runs the native W2 MMA and writes the output.

\paragraph{Accumulator-fold.}
When $M>1$, the bias-fold is not applicable because $R_m$ varies across rows.
We instead incorporate the row-dependent correction into the HMX accumulator, which we call the \emph{accumulator-fold}.
Since $R_m$ is the sum of multiple $q$-bit activation codes, the required correction $R_m/2$ can exceed the range of a single $q$-bit code.
We therefore decompose it as $R_m/2=R_c/2+L_m$, where $R_{c}$ is a per-call constant that can be folded into the per-channel input bias, and $L_{m}$ is a small row-dependent remainder.
To add $L_{m}$ to the accumulator, we append $S$ \emph{correction columns} $\Lambda$ whose row $m$ sums to $L_m$ to $X^q$ and $S$ corresponding columns of $+1$, a native W2 value, to $W^2$:
\begin{equation}
\begin{aligned}
  &P^{2}+L\,\mathbf{1}_N^{\top}
  =\bigl[\,X^q\;\big|\;\Lambda\,\bigr]
  \bigl[\,W^{2}\;\big|\;\mathbf{1}_{N\times S}\,\bigr]^{\top}, \\
  &Y^q_{m,n}\approx\Requant\bigl(P^{2}_{m,n}+L_m;\,2\alpha_n,\,
  (\widetilde b^{\mathrm{in}}_n-R_c)/2\bigr).
\end{aligned}
\label{eq:seq-augk}
\end{equation}
Before the first multiply-accumulate (MAC), HVX computes the row sums of the activation block and sets $R_{c}$ such that every $L_{m}$ is non-negative.
HVX then chooses the smallest $S$ for which each $L_{m}$ can be represented as the sum of $S$ $q$-bit codes, writes $\Lambda$, and folds $R_{c}$ into the input bias (see \Cref{app:augk}).
HMX then accumulates over the resulting $K+S$ columns and writes the final output using a single $\Requant$.
The only approximation occurs when $R_{m} - R_{c}$ is odd and the division by two must be rounded down.

\subsection{W3 via a Bit-Plane Fold}
\label{sec:hybrid-w3}

\paragraph{Representation.}
Each W3 weight is split into a signed W2 plane and an unsigned W1 plane,
so that $W^3=2W^2+W^1$ exactly:
\begin{equation}
  W^2_{n,k}=\left\lfloor W^3_{n,k}/2\right\rfloor,\qquad
  W^1_{n,k}=W^3_{n,k}-2W^2_{n,k}.
  \label{eq:w3-decomposition}
\end{equation}
Storing the two planes back-to-back requires three bits per weight physically without padding (see \Cref{app:w3-packing}).

\paragraph{Activation-fold.}
By linearity, $P^{3}=2P^{2}+P^{1}$.
A naive solution would requantize the two plane products separately, add the resulting outputs, and saturate the sum again, as in the output-correction form of W2-SEQ (see \Cref{app:output-correction-form}).
However, this approach requires an additional output pass and may lose precision because of repeated saturation.
We therefore propose the \emph{activation-fold}, which folds the factor of two into the activation rather than the output, allowing both plane products to share a single accumulator:
\begin{equation}
  \begin{aligned}
  \hat P^3_{m,n}&=P^{2}_{m,n}+\hat P^1_{m,n}
               =\frac{P^{3}_{m,n}-\delta_{m,n}}{2},\\
  \hat P^1_{m,n}&=\sum_k W^1_{n,k}\left\lfloor X^q_{m,k}/2\right\rfloor, \\
  \delta_{m,n}&=\sum_k W^1_{n,k}\,(X^q_{m,k}\bmod 2).
  \end{aligned}
  \label{eq:w3-fused-accumulator}
\end{equation}
Applying $\Requant$ with a doubled scale then yields
\begin{equation}
  \begin{aligned}
  Y^q_{m,n}&\approx\Requant\bigl(\hat P^3_{m,n};\,2\alpha_n,\,\hat b^{\mathrm{in},3}_n\bigr),\\
  \hat b^{\mathrm{in},3}_n&=\tfrac12 z_a\sum_k W^3_{n,k}-\tfrac14\sum_k W^1_{n,k},
  \end{aligned}
  \label{eq:w3-product-rewrite}
\end{equation}
where the first term in $\hat b^{\mathrm{in},3}_n$ is half of the W3 input bias defined in~\eqref{eq:bg-conversion-parameters} and the second term compensates for the expected value of $\delta_{m,n}/2$, assuming that odd and even activation codes are equally likely.
At runtime, the vector unit halves each activation once per call.
Since the output is requantized and saturated only once, the activation-fold avoids the precision loss caused by repeated saturation.
Its main approximation is dropping the activation LSB in the W1-plane product, and the second term in~\eqref{eq:w3-product-rewrite} compensates for its expected error, leaving at most half an LSB of error per activation code (see \Cref{app:fidelity}).

\paragraph{Parallel execution of the activation-fold.}
Although the activation-fold uses native low-bit operations, HMX must compute both $P^{2}_{m, n}$ and $\hat P^1_{m,n}$, resulting in $2MNK$ MACs.
At decode size, weight movement hides most, but not all, of this additional computation.
We therefore compute $\hat P^1_{1,n}$ on HVX at $M=1$, in parallel with the W2-plane MMA on HMX.
We call this the \emph{HVX-assisted} activation-fold and refer to the variant that computes both planes on HMX as the \emph{HMX-only} activation-fold.

Since the W1 plane is binary, HVX replaces multiplications with lookup-table (LUT) operations~\citep{wei2025tmac,wei2025tman}.
Concretely, HVX takes four activations $\lfloor X^q_{1,k}/2\rfloor$ and precomputes all $2^4$ subset sums.
The four W1 weights then index this table.
The table is built once per call and shared across all $N$ output channels, reducing the computation for each channel from $K$ MACs to $K/4$ lookups.
The resulting W1-plane products are added to the shared accumulator through extra tiles, like $\Lambda$ in~\eqref{eq:seq-augk}, allowing a single $\Requant{}$ to produce the final output.
Since the HVX computation cost grows with $M$, we compare the HVX-assisted and HMX-only activation-folds across $M$ in \Cref{app:hvx-assisted-vs-hmx-only}.

\subsection{Pipelined Execution and Path Selection}
\label{sec:lowbit-scheduling}

\paragraph{Pipelined execution.}
For each activation-row block, the kernel streams its weights in chunks of output-channel tiles.
DMA prefetches up to three chunks ahead into ring slots in VTCM, and HMX multiplies a chunk once it is ready.
HVX prepares the activation layout and materializes each chunk's outputs.
It also converts the weight chunks in the \naivepath{} and computes the intermediate terms of the \proposepath{}s.

\paragraph{From HVX-bound to memory-bound.}
\Cref{fig:lowbit-schedule} shows how the three units overlap at decode size ($M=1$).
In the \naivepath{}, every chunk must pass through HVX before HMX can use it.
Converting a chunk takes about $1.9\times$ its DMA time for W3 and $1.4\times$ for W2-SEQ.
HMX idles between chunks, and DMA stalls because a ring slot is freed only after its chunk is converted and multiplied.
The \proposepath{}s remove this stage, so DMA feeds encoded low-bit tiles directly to HMX.
For W3, HMX multiplies both planes into one accumulator (HMX-only).
For W2-SEQ, HVX computes the row sum and updates the input bias once, before the first MAC.
Both kernels are thus weight-stream-bound, like native kernels (\Cref{tab:bg-bandwidth}).
The fold paths reduce W3 and W2-SEQ latencies from 36.2 to 18.4~\textmu s and from 20.1 to 12.2~\textmu s, respectively.
The \naivepath{} reaches the weight-stream bound only with multiple HVX threads and large layers, as in Qwen3-4B (\Cref{app:upcast}).

\paragraph{Cost across $M$ and path selection.}
\Cref{tab:work-placement} generalizes this comparison to any $M$.
The \naivepath{} converts all $NK$ weights on HVX in every call, regardless of $M$.
The \proposepath{}s never convert weights.
In exchange, HMX does more MACs than the $MNK$ of a native kernel.
The W2-SEQ accumulator-fold and the W3 HVX-assisted activation-fold performs $MN(K{+}S)$ MACs, where one extra tile ($S=T$) suffices in practice, so the overhead is $T/K$.
Even the $2MNK$ of the W3 HMX-only activation-fold is cheap, since HMX delivers two to three orders of magnitude more throughput than an HVX thread~\citep{hao2026scaling}.
The trade therefore favors the \proposepath{}s at small $M$.
As $M$ grows, the $NK$ conversion is amortized over more rows, while the activation-side work and the extra MACs keep growing with $M$.
The crossover therefore depends on the MMA shape, motivating shape-dependent path selection in \sysname{} (\Cref{fig:paths-vs-m}).
\par

\section{Output-Channel Group-Wise MP}
\label{sec:group-mp}

\subsection{Hardware-Aligned MP Granularity}

HMX multiplies each weight tile in a single datatype (\Cref{fig:bg-hmx-architecture}). A weight tile spans $T$ contiguous output channels. Tiles along $K$ add into the same accumulator (see~\eqref{eq:bg-raw-accumulation}), so in principle they could use different bitwidths. However, all
tiles of an output channel share one weight scale (see~\eqref{eq:bg-conversion-parameters}).
Sharing a scale across tiles of different bitwidths prevents each bitwidth from using an independently calibrated quantization grid, which can introduce large quantization error.

We therefore assign bitwidths along the output channels. The finest granularity that keeps the native tiled layout is a group of $T$ contiguous output channels. Each group of a layer independently selects W2-SEQ, W3, or W4. All tiles of a group share its bitwidth, and each output channel keeps one scale across $K$. This meets both requirements of \Cref{sec:bg-challenges}.

\subsection{Profile Search}

We select the bitwidth of each output-channel group by minimizing the estimated quantization-induced loss increase under a target BPW budget.
We first estimate the cost of assigning each candidate bitwidth to each group, analyze how this cost varies within a layer, and then solve the resulting allocation problem using an integer linear program (ILP).

\begin{figure}[t]
  \centering
  \includegraphics[width=\linewidth]{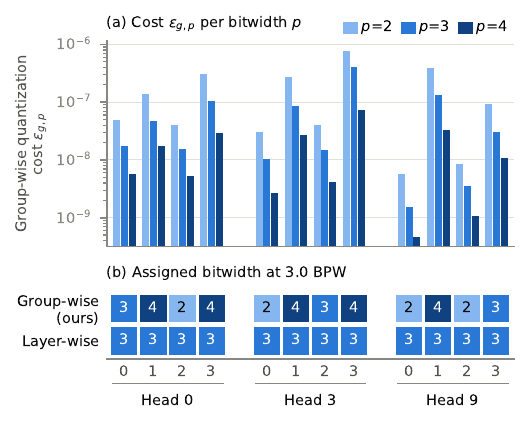}
  \vspace{-.7cm}
  \caption{Group-wise quantization sensitivity and bitwidth assignment for the query projection in the 12th Transformer block of Qwen3-4B (\texttt{model.layers.11.self\_attn.q\_proj}). (a) Quantization cost $\epsilon_{g, p}$ varies substantially across hardware-aligned output-channel groups. (b) Under a 3.0-BPW budget, the layer-wise profile assigns W3 uniformly, whereas the group-wise profile allocates different bitwidths according to group sensitivity.}
  \label{fig:group_sensitivity}
\end{figure}

\paragraph{Group-wise quantization cost}
Following prior Hessian-aware quantization methods (e.g., HAWQ-V2~\citep{dong2020hawq}), we use a second-order approximation to estimate the task loss increase induced by quantization.\footnote{$\Delta \mathcal{L} \approx \frac{1}{2} \Delta \mathbf{w}^{\top} \mathbf{H} \Delta \mathbf{w}$ where $\mathbf{H}$ is the Hessian with respect to the full-precision weights. In the sequel, we omit the common factor $\frac{1}{2}$, as it does not affect the resulting bitwidth assignment.}
Following SqueezeLLM~\citep{kim2024squeezellm}, we approximate the Hessian with the diagonal empirical Fisher, yielding the cost of assigning bitwidth $p$ to output-channel group $g$:
\begin{equation}
  \epsilon_{g,p}=\sum_{(n,k)\in g}
  \mathbb{E}_{\mathcal{D}_{\text{cal}}} \left [ \left ( \frac{\partial \mathcal{L}}{\partial W_{n, k}} \right )^{2} \right ] \left ( \Delta W_{n, k}^{p} \right )^{2},
\end{equation}
where $\mathcal{L}$ is the task loss, $\mathcal{D}_{\text{cal}}$ is the calibration set, and $\Delta W_{n,k}^{p} = s_{w,n}^{p} W_{n,k}^{p}-W_{n,k}$ is the perturbation introduced by quantizing $W_{n,k}$ to $p$-bit.
This cost jointly captures the sensitivity of each weight to the task loss and the magnitude of the perturbation introduced by the candidate bitwidth.

\Cref{fig:group_sensitivity}(a) shows the quantization costs $\epsilon_{g, p}$ of hardware-aligned groups from several attention heads within the same layer.
Although lower bitwidths generally incur higher costs, the absolute cost at each bitwidth varies substantially across groups.
Some groups exhibit low quantization costs at W2-SEQ, whereas others incur high costs even at W4.
A uniform bitwidth per layer may thus over-allocate precision to groups with low quantization costs or under-allocate it to groups with high ones.

\paragraph{Bitwidth allocation}
Let $\mathcal{P}$ be the \emph{palette}, the set of candidate formats, e.g., \{W2-SEQ, W3, W4\}, and let $b_p$ be the bitwidth of format $p\in\mathcal{P}$.
Let $x_{g,p}\in\{0,1\}$ indicate whether group $g$ is assigned format $p$, and $n_g$ the number of weights in group $g$.
Given a target BPW budget $B$, we solve
\begin{equation}
\begin{split}
\min_{x}\quad&
  \sum_{g,p} \epsilon_{g,p} x_{g,p} \\
\text{s.t.}\quad&
  \sum_{g,p} p\, n_g x_{g,p} \leq B \sum_{g} n_g,\qquad
  \sum_{p} x_{g,p}=1\;\;\forall g.
\end{split}
\end{equation}
The first constraint enforces the target average BPW, while the second assigns exactly one format to each output-channel group.
Solving the ILP yields a bitwidth assignment that minimizes the additive second-order loss surrogate under the specified BPW budget.
Because the \proposepath{}s introduced in \Cref{sec:kernels} run W2-SEQ and W3 as efficiently as native W4, the palette can include them, and groups of different formats are mixed within a single layer.

\Cref{fig:group_sensitivity}(b) illustrates the resulting bitwidth assignments under a 3.0-BPW budget. While the layer-wise profile uniformly assigns W3 to all groups, our group-wise profile assigns W2-SEQ to groups with lower quantization costs, W4 to groups with higher costs, and W3 to the remaining groups. This allocation better accounts for group-wise sensitivity variation under the same BPW budget.
\par

\section{Evaluation}
\label{sec:experiments}

We evaluate \sysname{} with three questions.
(Q1) Do the \proposepath{}s run W3 and W2-SEQ as close to the DMA ceiling as native kernels,
and up to which M (\Cref{sec:eval-kernels})?
(Q2) Do group-wise mixed-precision kernels scale with BPW in latency and energy (\Cref{sec:eval-mp-latency})?
(Q3) Does group-wise allocation improve accuracy over layer-wise allocation at the same BPW,
and do W2-SEQ and W3 add to it (\Cref{sec:eval-mp-acc})?
\Cref{sec:eval-e2e} then compares \sysname{} with llama.cpp end to end on the same NPU.

\subsection{Setup}
\label{sec:eval-setup}

\paragraph{Platform and models.}
We built a dedicated runtime for \sysname{} from scratch to implement fully static integer quantization, while reusing some operators, such as FlashAttention~\citep{dao2022flashattention}, from the llama.cpp Hexagon backend.
All latencies are measured on a mobile equipped with the \npuname{}.
We mainly use Qwen3 family models for evaluation: Qwen3-0.6B, 1.7B, 4B, and 8B (4B is an instruct model).
Kernel evaluations cover four projection shapes from Qwen3 decoder layers: QKV, O, Up/Gate, and Down projections.
We quantize the models using integer 16-bit activations (A16) and per-channel integer weights: uniform precision W4-, W3-, W2-, and W2-SEQ-A16, as well as mixed-precision profiles (MP-A16) with an output-channel-group size of $T=32$.
We also apply PrefixQuant~\citep{chen2026prefixquant} to reduce the activation quantization error caused by the wide activation ranges of prefix tokens (\Cref{app:eval-details}).
FlashAttention is excluded from integer quantization and uses an FP16 KV cache.

\paragraph{End-to-end metrics.}
We report decode (tg128, 128 generated tokens)
throughput in tokens per second on the NPU.
Accuracy of the quantized models is measured using GPU simulation with Qualcomm AIMET~\citep{siddegowda2022neuralnetworkquantizationai}, because running the full benchmark suite on the NPU is prohibitively time-consuming.
We validate the simulation by comparing its WikiText-2~\citep{merity2016pointer} perplexity (PPL) with on-target runs on the NPU (\Cref{app:ontarget-ppl}).
Using lm-evaluation-harness~\citep{eval-harness}, we report zero-shot commonsense reasoning (CSR) accuracy averaged over ARC-Challenge, ARC-Easy, HellaSwag, OpenBookQA, PIQA, and WinoGrande~\citep{allenai:arc,zellers2019hellaswag,OpenBookQA2018,Bisk2020,ai2:winogrande}.
We also report MMLU accuracy~\citep{hendrycks2020measuring} and IFEval~\citep{zhou2023instructionfollowingevaluationlargelanguage}.

\subsection{Non-Native Kernel Latency}
\label{sec:eval-kernels}

\begin{table}[t]
\centering
\caption{Qwen3 aggregate (QKV+O+Up/Gate+Down) decode-size ($M=1$) latency (\textmu s), and the effective weight bandwidth as \% of the DMA ceiling in \Cref{tab:bg-bandwidth}. Bold marks the fold paths (bias-fold for W2-SEQ and activation-fold for W3).}
\label{tab:m1-agg}
\small\setlength{\tabcolsep}{4pt}
\begin{tabular}{@{}llrrrrrrrr@{}}
\toprule
 & & \multicolumn{2}{c}{0.6B} & \multicolumn{2}{c}{1.7B} & \multicolumn{2}{c}{4B} & \multicolumn{2}{c}{8B} \\
\cmidrule(lr){3-4}\cmidrule(lr){5-6}\cmidrule(lr){7-8}\cmidrule(lr){9-10}
Format & Path & \textmu s & \% & \textmu s & \% & \textmu s & \% & \textmu s & \% \\
\midrule
W2 & native & 69 & 72 & 193 & 79 & 372 & 82 & 668 & 86 \\
\midrule
W2-SEQ & expand & 104 & 47 & 276 & 55 & 526 & 58 & 976 & 59 \\
 & \textbf{fold} & 75 & 66 & 194 & 79 & 372 & 82 & 672 & 86 \\
\midrule
\midrule
W4 & native & 114 & 84 & 345 & 87 & 667 & 91 & 1247 & 92 \\
\midrule
W3 & expand & 185 & 39 & 532 & 43 & 1046 & 44 & 1995 & 43 \\
 & \textbf{fold} & 94 & 77 & 270 & 84 & 516 & 88 & 954 & 91 \\
\bottomrule
\end{tabular}

\end{table}

\paragraph{Decode kernel latency.}
\Cref{tab:m1-agg} reports the sum of the four projection latencies during decoding ($M=1$) across all evaluated combinations of weight formats and execution paths on the four Qwen3 models.
Overall, the proposed \proposepath{}-based execution, which avoids weight expansion, enables W2-SEQ and W3 to achieve decode efficiency comparable to their native counterparts.
Specifically, W2-SEQ with the \proposepath{} nearly matches the decode latency of native W2.
Their effective weight bandwidth relative to the DMA ceiling differs less than 1\% except Qwen3-0.6B, whose small weight stream does not hide the per-call correction.
In contrast, W2-SEQ with the \naivepath{}, in which the weights are expanded to W4 before the MMA, is $1.41$--$1.52\times$ slower than native W2.
Similarly, W3 with the \naivepath{} is $1.55$--$1.63\times$ slower than native W4, whereas the \proposepath{} reduces its latency to $0.77$--$0.82\times$ that of native W4, approximately in proportion to the reduction in weight bytes.
Against the DMA ceiling, the \proposepath{} keeps W3 within 3 points of W4 except Qwen3-0.6B (7 points on 0.6B), while the \naivepath{} falls to 39--59\%.
The same performance trends carry over to end-to-end decoding on tg128 (\Cref{app:e2e-ablation-decode}).

\begin{figure}[t]
  \centering
  \includegraphics[width=\linewidth]{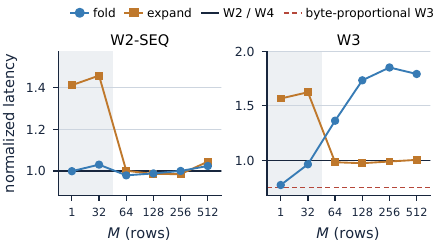}
  \caption{Qwen3-4B aggregate latency versus $M$: W2-SEQ normalized to W2 (left) and W3 normalized to W4 (right). Shaded: sizes where the \proposepath{} runs faster than the \naivepath{}. Dashed line: byte-proportional ideal of W3 (0.75 of W4).}
  \label{fig:paths-vs-m}
\end{figure}

\paragraph{Behaviour across $M$.}
We next examine the range of $M$ over which the \proposepath{} is beneficial, as its cost generally scales with $M$.
\Cref{fig:paths-vs-m} extends the preceding kernel-level analysis from decode to prefill sizes.
The \proposepath{} remains faster than the \naivepath{} for both W2-SEQ and W3 up to $M=32$ (see the other models in \Cref{fig:paths-vs-m-small}).
As $M$ increases, computation accounts for a larger fraction of latency relative to HMX MACs rather than weight movement.
Consequently, the relative latency of W3 deviates from the byte-proportional line because W3 requires twice as many HMX MACs as native W4.
Nevertheless, W3 remains faster than native W4 for $M \le 32$, except on Qwen3-0.6B, where the smaller weight stream limits the benefit of reducing weight traffic.
Based on this analysis, we configure the kernel dispatcher in \sysname{} to select an appropriate path for each MMA operation according to $M$.

\paragraph{Comparison with T-MAN.}
We reproduce T-MAN's kernels~\citep{wei2025tman} in \Cref{app:tman}.
On Qwen3-4B, W2-SEQ is $1.13\times$ faster at $M=1$, where both are bound by the weight stream, and $4.4$--$4.6\times$ faster for $8 \le M \le 32$, where T-MAN converts its bit-serial weights to FP16 before HMX.

\subsection{Mixed Precision Kernel Latency and Power}
\label{sec:eval-mp-latency}

\begin{figure}[t]
  \centering
  \includegraphics[width=\linewidth]{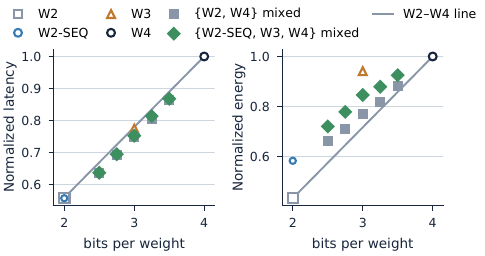}
  \caption{Mixed-precision kernels follow BPW in latency (left) and, with an offset for W2-SEQ and W3,
    in energy (right). Qwen3-4B, $M=1$, all projections aggregated, normalized to W4.
    Hollow markers are single-width kernels. The solid line interpolates W2 and W4.}
  \label{fig:e3-interpolation}
\end{figure}

At decode ($M=1$), the latency of an ideal mixed-precision kernel should scale with its BPW.
This requires that dispatching them across output-channel groups incurs negligible overhead.
\Cref{fig:e3-interpolation} tests this on Qwen3-4B with two palettes, \{W2, W4\} and \{W2-SEQ, W3, W4\}, from 2.5 to 3.5 BPW (see \Cref{app:e3-per-projection} for other models).
Each group contains $T=32$ output channels.
A marker placed on the interpolation line means the proper latency--BPW scaling.
We can find \sysname{} scales both \{W2, W4\} and \{W2-SEQ, W3, W4\} palettes well.
Furthermore, the \{W2-SEQ, W3, W4\} palette which includes non-native formats scales well enough like \{W2, W4\}.

Energy follows the interpolation line less closely in the right panel of \Cref{fig:e3-interpolation}, because \sysname{}'s W3 \proposepath{} spends extra computation on the W1-plane product.
W3 thus does not lie halfway between W2 and W4.
The same holds for W2-SEQ, which needs extra computation for folding.
This additional cost makes the \{W2-SEQ, W3, W4\} palette less energy-efficient than the pure native palette \{W2, W4\}.
Nevertheless, both palettes scale with BPW, even though their slopes deviate from the W2--W4 line.
\Cref{sec:eval-mp-acc} shows the accuracy this extra energy buys.

\subsection{Mixed Precision Model Accuracy}
\label{sec:eval-mp-acc}

\begin{figure}[t]
  \centering
  \includegraphics[width=\linewidth]{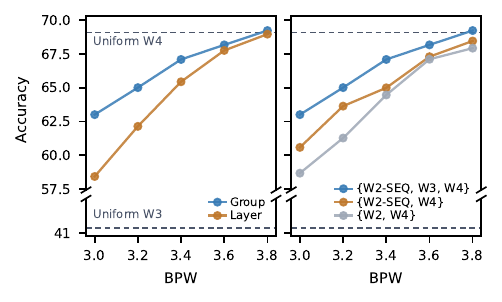}
  \vspace{-.5cm}
  \caption{Accuracy (average of CSR, MMLU, and IFEval) versus BPW of weight-only GPTQ MP profiles on Qwen3-4B. Left: output-channel-group-wise versus layer-wise allocation using the $\{$W2-SEQ, W3, W4$\}$ palette. Right: comparison of different bitwidth palettes using output-channel-group-wise allocation, where W2 denotes the conventional two's-complement format.}
  \label{fig:prelim-mp-result}
\end{figure}

\Cref{fig:prelim-mp-result} demonstrates the importance of both allocation granularity and bitwidth palette using weight-only mixed-precision profiles of Qwen3-4B.
As shown in the left panel, finer-grained allocation pushes the accuracy--BPW Pareto frontier outward.
Specifically, output-channel-group-wise allocation consistently outperforms layer-wise allocation, with a larger accuracy gain at lower BPW.
Enriching the bitwidth palette further improves the accuracy--BPW Pareto frontier (see the right panel).
Replacing W2 with the symmetric W2-SEQ format improves accuracy at the same BPW, and adding W3 provides an additional gain.
The gain from adding W2-SEQ to the palette grows at lower BPW, as ParetoQ reports~\citep{liu2026paretoq}.
The \proposepath{}s of \Cref{sec:kernels} let these non-native formats improve accuracy at a given BPW without sacrificing decode throughput. 

\subsection{End-to-End Accuracy and Efficiency}
\label{sec:eval-e2e}

\begin{figure*}[t]
  \centering
  \includegraphics[width=\linewidth]{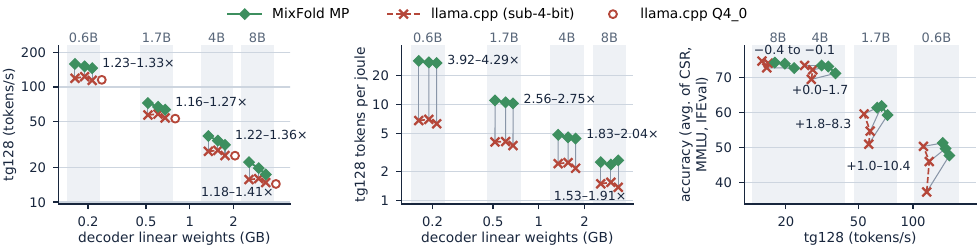}
  \caption{
  End-to-end decode throughput (Left), energy efficiency (Middle), and accuracy--throughput optimality (Right). Pairs matching BPW between \sysname{} and llama.cpp are connected in a line.
  }
  \label{fig:eval-pareto}
\end{figure*}

\paragraph{Baseline.}
We evaluate Q2\_K, Q3\_K\_S, Q3\_K\_M, and Q4\_0 using the llama.cpp Hexagon backend (build b11444), where the first three are sub-4-bit formats and Q4\_0 is a near-4-bit format.
We match each MP-A16 profile's target BPW to the corresponding sub-4-bit llama.cpp baseline.

\paragraph{Protocol.}
Following $\operatorname{llama-bench}$ practice~\citep{llamacpp}, we repeat five thermally controlled runs and report the median (see \Cref{app:e2e-protocol} for details).

\paragraph{Decode throughput.}
As shown in the left panel of \Cref{fig:eval-pareto}, \sysname{} achieves higher decode throughput than the corresponding llama.cpp baselines at matched BPW across all model sizes.
Moreover, the decode throughput of \sysname{} increases as BPW decreases.
This contrasts with llama.cpp, where sub-4-bit formats offer only limited speedups over Q4\_0.
For example, Q2\_K uses only 65\% of the weight storage of Q4\_0 but achieves just $1.03$--$1.09\times$ the decode throughput of Q4\_0.

\paragraph{Energy efficiency.}
\Cref{fig:eval-pareto} (middle) compares decode energy efficiency during tg128, measured in tokens per joule (J): a higher value indicates lower energy consumption per generated token.
We measure the power consumption of the entire system, including the CPU, NPU, and memory, throughout each run. 
At matched BPW, \sysname{} achieves $1.5$--$4.3\times$ higher energy efficiency than llama.cpp.

\paragraph{Accuracy--throughput optimality.}
For this evaluation, we quantize the weights according to each MP profile using FlexRound~\cite{lee2024flexroundlearnableroundingbased}, replacing the GPTQ-based quantization used in \Cref{fig:prelim-mp-result}.
FlexRound helps compensate for the accuracy loss caused by the coarser per-channel quantization used by \sysname{}, compared with the group-wise quantization used by llama.cpp.
As shown in the right panel of \Cref{fig:eval-pareto}, \sysname{} achieves competitive model accuracy to llama.cpp at matched BPW (from -0.4 to +10.4).
Moreover, \sysname{} achieves accuracy--throughput Pareto optimality.
These results suggest that, despite using coarser per-channel quantization, \sysname{} can achieve comparable or higher accuracy to llama.cpp at matched BPW by combining fine-grained mixed precision with an appropriate PTQ method, while maintaining high runtime efficiency.
\par

\section{Related Work}
\label{sec:related}

\paragraph{Executing low-bit weights.}
On GPUs, weight-only quantizers such as GPTQ and AWQ dequantize weights to
FP16 before the GEMM~\citep{frantar2022gptq,lin2024awq}, and later kernels hide that cost
behind the tensor cores~\citep{frantar2025marlin,xia2024fp6llm,zeng2025abq} or fold it into table lookups~\citep{park2024lut,guo2024flute}.
Atom and QServe keep the arithmetic in integers but at the bitwidths the tensor
cores provide~\citep{zhao2024atom,lin2025qserve}.
On CPUs, T-MAC and bitnet.cpp replace
the multiplication with table lookups~\citep{wei2025tmac,wang2025bitnet}.
On mobile NPUs, the llama.cpp Hexagon backend converts packed weights
to FP16 before the MMA~\citep{llamacpp}.
llm.npu keeps the NPU on its native INT8 path for prefill and runs decode
on the CPU~\citep{xu2025llmnpu}, and PowerInfer-2 splits decode between
the NPU and the CPU by activation density~\citep{xue2024powerinfer}.
MobileQuant and Quant.npu adapt the quantizer to the NPU's integer
arithmetic but stay at 4 bits or above~\citep{tan2024mobilequant,zhang2026quantnpu}.
T-MAN multiplies low-bit weights by table lookup on the NPU's vector cores,
which frees it from native datatypes but leaves the matrix unit idle at decode~\citep{wei2025tman}.
We compare its kernel with ours in \Cref{sec:eval-kernels}.
\citet{hao2026scaling} expose the NPU's tile shape to the quantizer.
ParetoQ shows that the symmetric W2-SEQ grid is more accurate than the
two's-complement grid, but gives no way to execute it~\citep{liu2026paretoq}.
\sysname{} neither converts weights nor leaves the matrix unit idle:
W2-SEQ and W3 run as native integer MMA, and the vector unit never converts weights at decode.

\paragraph{Mixed-precision allocation.}
Mixed precision trades allocation granularity against hardware fit.
HAWQ and its successors assign a bitwidth per layer from Hessian
sensitivity, and HAWQ-V3 solves the assignment as an
ILP~\citep{dong2019hawq,dong2020hawq,yao2021hawqv3}.
HAQ learns the assignment from measured latency~\citep{wang2019haq},
and differentiable relaxations search it by gradient~\citep{wu2018mixeddnas,yang2021fracbits};
both decide per layer.
For LLMs, SqueezeLLM, SpQR, and OWQ keep sensitive weights or columns at a
higher precision~\citep{kim2024squeezellm,dettmers2023spqr,lee2024owq},
SliM-LLM allocates per group by salience~\citep{huang2024slim},
MixLLM allocates per output channel~\citep{zheng2026mixllm},
and GAMMA learns a global allocation under a bit budget~\citep{yao2026gamma}.
\sysname{} admits the non-native W2-SEQ and W3 into the palette and adopts hardware-aware finer granularity to move the accuracy--throughput frontier outward.
\par

\section{Conclusion}
\label{sec:conclusion}

We presented \sysname{}, which runs sub-4-bit and mixed-precision LLM weights
on a mobile NPU without native low-bit support.
It executes W2-SEQ and W3 on native integer MMA without converting the weights,
and combines them with hardware-aware output-channel group-wise mixed precision.
On the \npuname{}, \sysname{} moves the accuracy--throughput frontier beyond llama.cpp's sub-4-bit formats while improving power efficiency.
All these gains follow from nativeness: no weight is converted before the
matrix unit, so decode is bound by weight bytes and the arithmetic stays in integers.
Hexagon is a representative case rather than a special one.
Mobile NPUs keep few native datatypes to save area and power,
so sub-4-bit formats are non-native on most of them.
\sysname{} needs only native 2-bit integer MMA,
per-channel requantization with a bias set at run time,
and a vector unit that shares on-chip memory with the matrix unit.
NPUs with these features can adopt the same execution paths.
\par

\bibliography{main}

@article{dong2020hawq,
  title={{HAWQ-V2}: Hessian Aware trace-Weighted Quantization of Neural Networks},
  author={Dong, Zhen and Yao, Zhewei and Arfeen, Daiyaan and Gholami, Amir and Mahoney, Michael W and Keutzer, Kurt},
  journal={Advances in Neural Information Processing Systems},
  volume={33},
  pages={18518--18529},
  year={2020}
}

@inproceedings{hao2026scaling,
  title={Scaling {LLM} Test-Time Compute with Mobile {NPU} on Smartphones},
  author={Hao, Zixu and Wei, Jianyu and Wang, Tuowei and Huang, Minxing and Jiang, Huiqiang and Jiang, Shiqi and Cao, Ting and Ren, Ju},
  booktitle={Proceedings of the 21st European Conference on Computer Systems},
  pages={2157--2172},
  year={2026}
}

@manual{qualcomm_hvx_v79,
  title        = {Qualcomm Hexagon V79 HVX Programmer's Reference Manual},
  author       = {{Qualcomm Technologies, Inc.}},
  organization = {Qualcomm Technologies, Inc.},
  number       = {80-N2040-61},
  year         = {2025},
  url          = {https://docs.qualcomm.com/doc/80-N2040-61/}
}

@manual{qualcomm_hmx_v81,
  title        = {Qualcomm Hexagon V81 HMX Programmer's Reference Manual},
  author       = {{Qualcomm Technologies, Inc.}},
  organization = {Qualcomm Technologies, Inc.},
  number       = {80-N2040-62},
  edition      = {Rev. AA},
  year         = {2026},
  month        = mar,
  url          = {https://docs.qualcomm.com/doc/80-N2040-62/80-N2040-62_REV_AA_Qualcomm_Hexagon_V81_HMX_Programmers_Reference_Manual.pdf}
}

@article{nagel2021white,
  title={A White Paper on Neural Network Quantization},
  author={Nagel, Markus and Fournarakis, Marios and Amjad, Rana Ali and Bondarenko, Yelysei and Van Baalen, Mart and Blankevoort, Tijmen},
  journal={arXiv preprint arXiv:2106.08295},
  year={2021}
}

@inproceedings{noh2025jack,
  title={Jack Unit: An Area- and Energy-Efficient Multiply-Accumulate ({MAC}) Unit Supporting Diverse Data Formats},
  author={Noh, Seock-Hwan and Kim, Sungju and Kim, Seohyun and Kim, Daehoon and Kung, Jaeha and Kim, Yeseong},
  booktitle={2025 IEEE/ACM International Symposium on Low Power Electronics and Design (ISLPED)},
  pages={1--7},
  year={2025},
  organization={IEEE}
}

@inproceedings{horowitz20141,
  title={1.1 Computing's Energy Problem (and What We Can Do About It)},
  author={Horowitz, Mark},
  booktitle={2014 IEEE international solid-state circuits conference digest of technical papers (ISSCC)},
  pages={10--14},
  year={2014},
  organization={Ieee}
}

@article{park2022multi,
title = "A Multi-Mode 8k-MAC HW-Utilization-Aware Neural Processing Unit with a Unified Multi-Precision Datapath in 4-nm Flagship Mobile SoC",
author = "Park, \{Jun Seok\} and Changsoo Park and Suknam Kwon and Taeho Jeon and Yesung Kang and Heonsoo Lee and Dongwoo Lee and James Kim and Kim, \{Hyeong Seok\} and Lee, \{Young Jong\} and Sangkyu Park and Kim, \{Min Seong\} and Ha, \{Sang Hyuck\} and Jihoon Bang and Jinpyo Park and Lim, \{Suk Hwan\} and Inyup Kang",
note = "Publisher Copyright: {\textcopyright} 2022 IEEE.",
year = "2023",
month = jan,
day = "1",
doi = "10.1109/JSSC.2022.3205713",
language = "English",
volume = "58",
pages = "189--202",
journal = "IEEE Journal of Solid-State Circuits",
issn = "0018-9200",
publisher = "Institute of Electrical and Electronics Engineers Inc.",
number = "1",
}

@misc{llamacpp,
  author       = {{The ggml authors}},
  title        = {{llama.cpp}: {LLM} Inference in {C/C++}},
  howpublished = {\url{https://github.com/ggml-org/llama.cpp}},
  year         = {2023},
  note         = {GitHub repository}
}

@INPROCEEDINGS{mahurin2023qualcomm,
  author={Mahurin, Eric},
  booktitle={2023 IEEE Hot Chips 35 Symposium (HCS)}, 
  title={{Qualcomm Hexagon NPU}}, 
  year={2023},
  volume={},
  number={},
  pages={1-19},
  doi={10.1109/HCS59251.2023.10254715}}

@article{chen2026prefixquant,
  title={{PrefixQuant}: Eliminating Outliers by Prefixed Tokens for Large Language Models Quantization},
  author={Chen, Mengzhao and Liu, Yi and Wang, Jiahao and Bin, Yi and Shao, Wenqi and Luo, Ping},
  journal={IEEE Transactions on Pattern Analysis and Machine Intelligence},
  year={2026},
  publisher={IEEE}
}

@inproceedings{liu2026paretoq,
 author = {Liu, Zechun and Zhao, Changsheng and Huang, Hanxian and Chen, Sijia and Zhang, Jing and Zhao, Jiawei and Roy, Scott and Jin, Lisa and Xiong, Yunyang and Shi, Yangyang and Xiao, Lin and Tian, Yuandong and Soran, Bilge and Krishnamoorthi, Raghuraman and Blankevoort, Tijmen and Chandra, Vikas},
 booktitle = {Advances in Neural Information Processing Systems},
 doi = {10.52202/085713-3055},
 editor = {D. Belgrave and C. Zhang and H. Lin and R. Pascanu and P. Koniusz and M. Ghassemi and N. Chen},
 pages = {91311--91336},
 publisher = {Curran Associates, Inc.},
 title = {ParetoQ: Improving Scaling Laws in Extremely Low-bit LLM Quantization},
 volume = {38, Main Conference},
 year = {2025}
}

@article{yao2026gamma,
  title={{GAMMA}: Global Bit Allocation for Mixed-Precision Models under Arbitrary Budgets},
  author={Yao, Zhangyang and Zhao, Haiyan and Wang, Haoyu and Huang, Tianbo and Zhang, Lihua and Han, Xu},
  journal={arXiv preprint arXiv:2605.18475},
  year={2026}
}

@article{zheng2026mixllm,
  title={{MixLLM}: {LLM} Quantization with Global Mixed-Precision between Output-Features and Highly-Efficient System Design},
  author={Zheng, Zhen and Song, Xiaonan and Liu, Chuanjie},
  journal={Proceedings of Machine Learning and Systems},
  volume={8},
  pages={1771--1785},
  year={2026}
}

@misc{siddegowda2022neuralnetworkquantizationai,
      title={Neural Network Quantization with AI Model Efficiency Toolkit (AIMET)}, 
      author={Sangeetha Siddegowda and Marios Fournarakis and Markus Nagel and Tijmen Blankevoort and Chirag Patel and Abhijit Khobare},
      year={2022},
      eprint={2201.08442},
      archivePrefix={arXiv},
      primaryClass={cs.LG},
      url={https://arxiv.org/abs/2201.08442}, 
}

@inproceedings{kim2024squeezellm,
  title={{SqueezeLLM}: Dense-and-Sparse Quantization},
  author={Kim, Sehoon and Hooper, Coleman and Gholami, Amir and Dong, Zhen and Li, Xiuyu and Shen, Sheng and Mahoney, Michael W. and Keutzer, Kurt},
  booktitle={International Conference on Machine Learning},
  year={2024}
}

@misc{eval-harness,
  author       = {Gao, Leo and Tow, Jonathan and Abbasi, Baber and Biderman, Stella and Black, Sid and DiPofi, Anthony and Foster, Charles and Golding, Laurence and Hsu, Jeffrey and Le Noac'h, Alain and Li, Haonan and McDonell, Kyle and Muennighoff, Niklas and Ociepa, Chris and Phang, Jason and Reynolds, Laria and Schoelkopf, Hailey and Skowron, Aviya and Sutawika, Lintang and Tang, Eric and Thite, Anish and Wang, Ben and Wang, Kevin and Zou, Andy},
  title        = {The Language Model Evaluation Harness},
  month        = 07,
  year         = 2024,
  publisher    = {Zenodo},
  version      = {v0.4.3},
  doi          = {10.5281/zenodo.12608602},
  url          = {https://zenodo.org/records/12608602}
}

@article{allenai:arc,
      author    = {Peter Clark  and Isaac Cowhey and Oren Etzioni and Tushar Khot and
                    Ashish Sabharwal and Carissa Schoenick and Oyvind Tafjord},
      title     = {Think you have Solved Question Answering? Try ARC, the AI2 Reasoning Challenge},
      journal   = {arXiv:1803.05457v1},
      year      = {2018},
}

@inproceedings{zellers2019hellaswag,
    title={HellaSwag: Can a Machine Really Finish Your Sentence?},
    author={Zellers, Rowan and Holtzman, Ari and Bisk, Yonatan and Farhadi, Ali and Choi, Yejin},
    booktitle ={Proceedings of the 57th Annual Meeting of the Association for Computational Linguistics},
    year={2019}
}

@inproceedings{Bisk2020,
  author = {Yonatan Bisk and Rowan Zellers and
            Ronan Le Bras and Jianfeng Gao
            and Yejin Choi},
  title = {PIQA: Reasoning about Physical Commonsense in
           Natural Language},
  booktitle = {Thirty-Fourth AAAI Conference on
               Artificial Intelligence},
  year = {2020},
}

@article{ai2:winogrande,
    title={WinoGrande: An Adversarial Winograd Schema Challenge at Scale},
    author={Sakaguchi, Keisuke and Bras, Ronan Le and Bhagavatula, Chandra and Choi, Yejin},
    journal={arXiv preprint arXiv:1907.10641},
    year={2019}
}

@inproceedings{OpenBookQA2018,
 title={Can a Suit of Armor Conduct Electricity? A New Dataset for Open Book Question Answering},
 author={Todor Mihaylov and Peter Clark and Tushar Khot and Ashish Sabharwal},
 booktitle={EMNLP},
 year={2018}
}

@article{hendrycks2020measuring,
  title={Measuring massive multitask language understanding},
  author={Hendrycks, Dan and Burns, Collin and Basart, Steven and Zou, Andy and Mazeika, Mantas and Song, Dawn and Steinhardt, Jacob},
  journal={arXiv preprint arXiv:2009.03300},
  year={2020}
}

@misc{zhou2023instructionfollowingevaluationlargelanguage,
      title={Instruction-Following Evaluation for Large Language Models}, 
      author={Jeffrey Zhou and Tianjian Lu and Swaroop Mishra and Siddhartha Brahma and Sujoy Basu and Yi Luan and Denny Zhou and Le Hou},
      year={2023},
      eprint={2311.07911},
      archivePrefix={arXiv},
      primaryClass={cs.CL},
      url={https://arxiv.org/abs/2311.07911}, 
}

@misc{merity2016pointer,
      title={Pointer Sentinel Mixture Models},
      author={Stephen Merity and Caiming Xiong and James Bradbury and Richard Socher},
      year={2016},
      eprint={1609.07843},
      archivePrefix={arXiv},
      primaryClass={cs.CL}
}

@inproceedings{zeng2025abq,
  title={Abq-llm: Arbitrary-bit quantized inference acceleration for large language models},
  author={Zeng, Chao and Liu, Songwei and Xie, Yusheng and Liu, Hong and Wang, Xiaojian and Wei, Miao and Yang, Shu and Chen, Fangmin and Mei, Xing},
  booktitle={Proceedings of the AAAI Conference on Artificial Intelligence},
  volume={39},
  number={21},
  pages={22299--22307},
  year={2025}
}

@article{wei2025tman,
  title={T-MAN: Enabling End-to-End Low-Bit LLM Inference on NPUs via Unified Table Lookup},
  author={Wei, Jianyu and Li, Qingtao and Cao, Shijie and Ma, Lingxiao and Hao, Zixu and Zhang, Yanyong and Hu, Xiaoyan and Cao, Ting},
  journal={arXiv preprint arXiv:2511.11248},
  year={2025}
}

@inproceedings{wei2025tmac,
  title={T-mac: Cpu renaissance via table lookup for low-bit llm deployment on edge},
  author={Wei, Jianyu and Cao, Shijie and Cao, Ting and Ma, Lingxiao and Wang, Lei and Zhang, Yanyong and Yang, Mao},
  booktitle={Proceedings of the Twentieth European Conference on Computer Systems},
  pages={278--292},
  year={2025}
}

@InProceedings{lee2024flexroundlearnableroundingbased,
  title = 	 {{F}lex{R}ound: Learnable Rounding based on Element-wise Division for Post-Training Quantization},
  author =       {Lee, Jung Hyun and Kim, Jeonghoon and Kwon, Se Jung and Lee, Dongsoo},
  booktitle = 	 {Proceedings of the 40th International Conference on Machine Learning},
  pages = 	 {18913--18939},
  year = 	 {2023},
  editor = 	 {Krause, Andreas and Brunskill, Emma and Cho, Kyunghyun and Engelhardt, Barbara and Sabato, Sivan and Scarlett, Jonathan},
  volume = 	 {202},
  series = 	 {Proceedings of Machine Learning Research},
  month = 	 {23--29 Jul},
  publisher =    {PMLR},
  url = 	 {https://proceedings.mlr.press/v202/lee23h.html}
}

@inproceedings{wang2025bitnet,
  title={Bitnet. cpp: Efficient edge inference for ternary llms},
  author={Wang, Jinheng and Zhou, Hansong and Song, Ting and Cao, Shijie and Xia, Yan and Cao, Ting and Wei, Jianyu and Ma, Shuming and Wang, Hongyu and Wei, Furu},
  booktitle={Proceedings of the 63rd Annual Meeting of the Association for Computational Linguistics (Volume 1: Long Papers)},
  pages={9305--9322},
  year={2025}
}

@article{frantar2022gptq,
  title={Gptq: Accurate post-training quantization for generative pre-trained transformers},
  author={Frantar, Elias and Ashkboos, Saleh and Hoefler, Torsten and Alistarh, Dan},
  journal={arXiv preprint arXiv:2210.17323},
  year={2022}
}

@article{lin2024awq,
  title={Awq: Activation-aware weight quantization for on-device llm compression and acceleration},
  author={Lin, Ji and Tang, Jiaming and Tang, Haotian and Yang, Shang and Chen, Wei-Ming and Wang, Wei-Chen and Xiao, Guangxuan and Dang, Xingyu and Gan, Chuang and Han, Song},
  journal={Proceedings of machine learning and systems},
  volume={6},
  pages={87--100},
  year={2024}
}

@inproceedings{frantar2025marlin,
  title={Marlin: Mixed-precision auto-regressive parallel inference on large language models},
  author={Frantar, Elias and Castro, Roberto L and Chen, Jiale and Hoefler, Torsten and Alistarh, Dan},
  booktitle={Proceedings of the 30th ACM SIGPLAN Annual Symposium on Principles and Practice of Parallel Programming},
  pages={239--251},
  year={2025}
}

@misc{xia2024fp6llm,
      title={FP6-LLM: Efficiently Serving Large Language Models Through FP6-Centric Algorithm-System Co-Design}, 
      author={Haojun Xia and Zhen Zheng and Xiaoxia Wu and Shiyang Chen and Zhewei Yao and Stephen Youn and Arash Bakhtiari and Michael Wyatt and Donglin Zhuang and Zhongzhu Zhou and Olatunji Ruwase and Yuxiong He and Shuaiwen Leon Song},
      year={2024},
      eprint={2401.14112},
      archivePrefix={arXiv},
      primaryClass={cs.LG},
      url={https://arxiv.org/abs/2401.14112}, 
}

@inproceedings{park2024lut,
 author = {Park, Gunho and park, baeseong and Kim, Minsub and Lee, Sungjae and Kim, Jeonghoon and Kwon, Beomseok and Kwon, Se Jung and Kim, Byeongwook and Lee, Youngjoo and Lee, Dongsoo},
 booktitle = {International Conference on Learning Representations},
 editor = {B. Kim and Y. Yue and S. Chaudhuri and K. Fragkiadaki and M. Khan and Y. Sun},
 pages = {38069--38086},
 title = {LUT-GEMM: Quantized Matrix Multiplication based on LUTs for Efficient Inference in Large-Scale Generative Language Models},
 volume = {2024},
 year = {2024}
}

@inproceedings{guo2024flute,
  title={Fast matrix multiplications for lookup table-quantized llms},
  author={Guo, Han and Brandon, William and Cholakov, Radostin and Ragan-Kelley, Jonathan and Xing, Eric and Kim, Yoon},
  booktitle={Findings of the Association for Computational Linguistics: EMNLP 2024},
  pages={12419--12433},
  year={2024}
}

@article{zhao2024atom,
  title={Atom: Low-bit quantization for efficient and accurate llm serving},
  author={Zhao, Yilong and Lin, Chien-Yu and Zhu, Kan and Ye, Zihao and Chen, Lequn and Zheng, Size and Ceze, Luis and Krishnamurthy, Arvind and Chen, Tianqi and Kasikci, Baris},
  journal={Proceedings of Machine Learning and Systems},
  volume={6},
  pages={196--209},
  year={2024}
}

@article{lin2025qserve,
  title={Qserve: W4a8kv4 quantization and system co-design for efficient llm serving},
  author={Lin, Yujun and Tang, Haotian and Yang, Shang and Zhang, Zhekai and Xiao, Guangxuan and Gan, Chuang and Han, Song},
  journal={Proceedings of Machine Learning and Systems},
  volume={7},
  year={2025}
}

@inproceedings{xu2025llmnpu,
  title={Fast on-device llm inference with npus},
  author={Xu, Daliang and Zhang, Hao and Yang, Liming and Liu, Ruiqi and Huang, Gang and Xu, Mengwei and Liu, Xuanzhe},
  booktitle={Proceedings of the 30th ACM International Conference on Architectural Support for Programming Languages and Operating Systems, Volume 1},
  pages={445--462},
  year={2025}
}

@article{xue2024powerinfer,
  title={Powerinfer-2: Fast large language model inference on a smartphone},
  author={Xue, Zhenliang and Song, Yixin and Mi, Zeyu and Zheng, Xinrui and Xia, Yubin and Chen, Haibo},
  journal={arXiv preprint arXiv:2406.06282},
  year={2024}
}

@inproceedings{tan2024mobilequant,
  title={Mobilequant: Mobile-friendly quantization for on-device language models},
  author={Tan, Fuwen and Lee, Royson and Dudziak, {\L}ukasz and Hu, Shell Xu and Bhattacharya, Sourav and Hospedales, Timothy and Tzimiropoulos, Georgios and Martinez, Brais},
  booktitle={Findings of the Association for Computational Linguistics: EMNLP 2024},
  pages={9761--9771},
  year={2024}
}

@article{zhang2026quantnpu,
  title={Quant. npu: Enabling Efficient Mobile NPU Inference for on-device LLMs via Fully Static Quantization},
  author={Zhang, Jinghe and Xu, Daliang and Wang, Chenghua and Xie, Weikai and Qi, Tao and Ma, Yun and Xu, Mengwei and Huang, Gang},
  journal={arXiv preprint arXiv:2605.20295},
  year={2026}
}

@InProceedings{yao2021hawqv3,
  title = 	 {HAWQ-V3: Dyadic Neural Network Quantization},
  author =       {Yao, Zhewei and Dong, Zhen and Zheng, Zhangcheng and Gholami, Amir and Yu, Jiali and Tan, Eric and Wang, Leyuan and Huang, Qijing and Wang, Yida and Mahoney, Michael and Keutzer, Kurt},
  booktitle = 	 {Proceedings of the 38th International Conference on Machine Learning},
  pages = 	 {11875--11886},
  year = 	 {2021},
  editor = 	 {Meila, Marina and Zhang, Tong},
  volume = 	 {139},
  series = 	 {Proceedings of Machine Learning Research},
  month = 	 {18--24 Jul},
  publisher =    {PMLR},
  url = 	 {https://proceedings.mlr.press/v139/yao21a.html}
}

@inproceedings{dong2019hawq,
  title={Hawq: Hessian aware quantization of neural networks with mixed-precision},
  author={Dong, Zhen and Yao, Zhewei and Gholami, Amir and Mahoney, Michael W and Keutzer, Kurt},
  booktitle={Proceedings of the IEEE/CVF international conference on computer vision},
  pages={293--302},
  year={2019}
}

@inproceedings{wang2019haq,
  title={Haq: Hardware-aware automated quantization with mixed precision},
  author={Wang, Kuan and Liu, Zhijian and Lin, Yujun and Lin, Ji and Han, Song},
  booktitle={2019 IEEE/CVF conference on computer vision and pattern recognition (CVPR)},
  pages={8604--8612},
  year={2019},
  organization={IEEE}
}

@article{wu2018mixeddnas,
  title={Mixed precision quantization of convnets via differentiable neural architecture search},
  author={Wu, Bichen and Wang, Yanghan and Zhang, Peizhao and Tian, Yuandong and Vajda, Peter and Keutzer, Kurt},
  journal={arXiv preprint arXiv:1812.00090},
  year={2018}
}

@inproceedings{yang2021fracbits,
  title={Fracbits: Mixed precision quantization via fractional bit-widths},
  author={Yang, Linjie and Jin, Qing},
  booktitle={Proceedings of the AAAI Conference on Artificial Intelligence},
  volume={35},
  number={12},
  pages={10612--10620},
  year={2021}
}

@article{dettmers2023spqr,
  title={Spqr: A sparse-quantized representation for near-lossless llm weight compression},
  author={Dettmers, Tim and Svirschevski, Ruslan and Egiazarian, Vage and Kuznedelev, Denis and Frantar, Elias and Ashkboos, Saleh and Borzunov, Alexander and Hoefler, Torsten and Alistarh, Dan},
  journal={arXiv preprint arXiv:2306.03078},
  year={2023}
}

@inproceedings{lee2024owq,
  title={Owq: Outlier-aware weight quantization for efficient fine-tuning and inference of large language models},
  author={Lee, Changhun and Jin, Jungyu and Kim, Taesu and Kim, Hyungjun and Park, Eunhyeok},
  booktitle={Proceedings of the AAAI Conference on Artificial Intelligence},
  volume={38},
  number={12},
  pages={13355--13364},
  year={2024}
}

@article{huang2024slim,
  title={SliM-LLM: Salience-driven mixed-precision quantization for large language models},
  author={Huang, Wei and Qin, Haotong and Liu, Yangdong and Li, Yawei and Liu, Qinshuo and Liu, Xianglong and Benini, Luca and Magno, Michele and Zhang, Shiming and Qi, Xiaojuan},
  journal={arXiv preprint arXiv:2405.14917},
  year={2024}
}

@article{dao2022flashattention,
  title={Flashattention: Fast and memory-efficient exact attention with io-awareness},
  author={Dao, Tri and Fu, Dan and Ermon, Stefano and Rudra, Atri and R{\'e}, Christopher},
  journal={Advances in neural information processing systems},
  volume={35},
  pages={16344--16359},
  year={2022}
}

@INPROCEEDINGS{lucian2015dspimaging,
  author={Codrescu, Lucian},
  booktitle={2015 IEEE Hot Chips 27 Symposium (HCS)}, 
  title={Architecture of the Hexagon™ 680 DSP for mobile imaging and computer vision}, 
  year={2015},
  volume={},
  number={},
  pages={1-26},
  doi={10.1109/HOTCHIPS.2015.7477329}}
\bibliographystyle{paperstyle}

\clearpage
\section*{AI Usage Disclosure}
Generative AI tools (Claude Code and Claude, Anthropic) were used in producing this work as follows.

\textbf{Code generation.}
An LLM-based coding assistant was used
extensively to implement and debug the runtime under the authors'
direction: the DSP-side HVX and HMX kernels, the host runtime, the unit
tests and the weight exporter. The authors specified the designs,
reviewed every change, and validated it with the bit-exact unit tests and
the on-device measurements reported in the paper.

\textbf{Experiments and data processing.}
The benchmark drivers, the result archive and the scripts that produce the
tables and figures from the archived measurements were written with the
same assistant. All measurements were taken by the authors on the device
described in \Cref{sec:eval-setup}. No experimental data were generated or
altered by an AI tool.

\textbf{Writing.}
The authors wrote the draft text, captions and tables from their notes and the code.
The LLM assistant was used to check the equations against the implementation, and to proofread the English.
The authors wrote the problem framing and the claims, reviewed the text, and are responsible for its content.
At the authors' request, the LLM assistant also drafted and rephrased sentences, mainly during revision.

\textbf{Ideation, experiment design, and literature search.}
The authors performed the ideation and experiment design without the use of AI tools.
Literature search was performed with the help of the LLM assistant which was used to identify relevant references.
The authors reviewed the references and selected the relevant ones for inclusion in the paper.

\clearpage
\appendix
\section{Limitations}
\label{app:arxiv-limitations}

\paragraph{Effectiveness in single-thread vector unit.}
The performance advantage of our fold methods is established against
single-thread weight expansion. In this setting, the vector unit converts
unsupported weight formats before the matrix unit can consume them, and
this conversion can lie on the critical path. Folding removes the expansion
stage and allows the reduced weight traffic to translate into lower
latency. However, processing throughput increase by multiple vector threads can
reduce the conversion time and narrow or eliminate this advantage.

\paragraph{Unverified end-to-end gains for multi-row decoding.}
Our end-to-end evaluation covers single-stream autoregressive decoding at
$M=1$. We have not verified the end-to-end effectiveness of \sysname{}
for $M>1$ decoding, including speculative decoding and concurrent decoding
of multiple sequences. Evaluating the fold paths and mixed-precision profiles within
these workloads remains future work.
\renewcommand{\topfraction}{0.9}
\renewcommand{\bottomfraction}{0.8}
\renewcommand{\textfraction}{0.07}
\renewcommand{\floatpagefraction}{0.8}
\renewcommand{\dbltopfraction}{0.9}
\renewcommand{\dblfloatpagefraction}{0.8}
\setcounter{topnumber}{4}
\setcounter{bottomnumber}{3}
\setcounter{totalnumber}{6}
\setcounter{dbltopnumber}{3}
\section{Implementation Details}
\label{app:implementation}

\subsection{Output-Correction Form}
\label{app:output-correction-form}

\Cref{eq:seq-converted-output} needs the row-dependent input bias
$(\widetilde b^{\mathrm{in}}_n-R_m)/2$. The output-correction form instead
requantizes with a reference row sum $\bar R$, fixed from the calibration
set, and adds the remainder after the conversion:
\begin{equation}
  \begin{aligned}
  Y^q_{m,n}&\approx\operatorname{clip}\bigl(\widehat Y_{m,n}+\lfloor C_{m,n}\rceil\bigr),\\
  \widehat Y_{m,n}&=\Requant\bigl(P^{2}_{m,n};\,2\alpha_n,\,
  (\widetilde b^{\mathrm{in}}_n-\bar R)/2\bigr),
  \end{aligned}
  \label{eq:seq-output-correction}
\end{equation}
with $C_{m,n}=\alpha_n(R_m-\bar R)$. HMX writes $\widehat Y_{m,n}$, and HVX computes $R_m$ from the activation tiles
and adds $C_{m,n}$ to all $M\times N$ outputs.

\paragraph{Intermediate clipping.}
$\widehat Y_{m,n}$ is saturated to $[0,2^q-1]$ before $C_{m,n}$ is added.
An output whose final value lies in the code range but whose intermediate
does not is therefore clipped, and the correction cannot recover it. The
bias-fold and the accumulator-fold apply the correction before HMX
saturates the output, so they saturate once and have no such clipping.

\subsection{W3 Plane Packing}
\label{app:w3-packing}

A W3 code with bits $c_2c_1c_0$ is stored as the raw
W2 code $(c_2\mathbin{\ll}1)\mathbin{|}c_1$ and the raw W1 code $c_0$.
The W2 raw code is interpreted as signed two's complement by the native HMX
kernel, so raw codes 0, 1, 2, and 3 represent 0, 1, $-2$, and $-1$,
respectively. The W1 plane is unsigned. HMX reads the W2 plane in its signed
2-bit weight mode and the W1 plane in its unsigned 1-bit weight mode, so
neither plane is widened. All W2 tiles are stored first and all W1 tiles
second. For $N$ and $K$ divisible by 32, the total
weight payload is
\begin{equation}
  \frac{N}{32}\frac{K}{32}(256+128)
  = \frac{3NK}{8}\ \text{bytes}.
\end{equation}

\subsection{Expand Baseline}
\label{app:upcast}

\begin{table*}[!htbp]
  \centering
  \caption{Decode latency with more HVX threads, over the fold without HVX
  weight work (Fold: the W2-SEQ bias- and accumulator-folds and the W3
  HMX-only activation-fold), summed over QKV, O, Up/Gate, and Down. $n$
  counts the HVX threads on weight work. Expand with $n=1$ converts on the
  issuing thread.}
  \label{tab:hvx-budget}
  \footnotesize
  \begin{tabular}{@{}llrlrrrrrrrr@{}}
\toprule
& & & & & \multicolumn{7}{c}{Latency over Fold, $n$ HVX threads} \\
\cmidrule(l){6-12}
Model & Format & $M$ & Path & Fold (\textmu s) & 1 & 2 & 3 & 4 & 5 & 6 & 7 \\
\midrule
0.6B & W2-SEQ & 1 & Expand & 75 & 1.41 & 1.13 & 1.07 & 1.05 & 1.04 & 1.04 & 1.04 \\
 &  & 8 & Expand & 83 & 1.41 & 1.15 & 1.09 & 1.08 & 1.06 & 1.07 & 1.06 \\
 &  & 32 & Expand & 116 & 1.28 & 1.37 & 1.34 & 1.32 & 1.32 & 1.32 & 1.32 \\
 & W3 & 1 & Expand & 108 & 1.72 & 1.09 & 1.01 & 0.97 & 0.95 & 0.93 & 0.93 \\
 &  & 1 & HVX-assisted fold & 108 & 1.39 & 0.93 & 0.88 & 0.88 & 0.87 & 0.87 & 0.87 \\
 &  & 8 & Expand & 121 & 1.66 & 1.06 & 0.97 & 0.94 & 0.92 & 0.91 & 0.91 \\
 &  & 32 & Expand & 160 & 1.45 & 1.17 & 1.11 & 1.09 & 1.07 & 1.07 & 1.06 \\
\midrule
4B & W2-SEQ & 1 & Expand & 372 & 1.42 & 1.03 & 1.00 & 0.98 & 0.97 & 0.97 & 0.97 \\
 &  & 8 & Expand & 390 & 1.41 & 1.03 & 1.00 & 0.98 & 0.98 & 0.98 & 0.98 \\
 &  & 32 & Expand & 437 & 1.43 & 1.10 & 1.06 & 1.05 & 1.04 & 1.04 & 1.05 \\
 & W3 & 1 & Expand & 577 & 1.81 & 1.04 & 0.94 & 0.91 & 0.90 & 0.90 & 0.90 \\
 &  & 1 & HVX-assisted fold & 577 & 1.40 & 0.92 & 0.90 & 0.90 & 0.90 & 0.89 & 0.89 \\
 &  & 8 & Expand & 593 & 1.82 & 1.04 & 0.94 & 0.92 & 0.91 & 0.90 & 0.90 \\
 &  & 32 & Expand & 685 & 1.70 & 1.00 & 0.91 & 0.90 & 0.89 & 0.88 & 0.88 \\
\bottomrule
\end{tabular}

\end{table*}

\Cref{alg:naive-w4-upcast} gives the expand path used as the baseline in
\Cref{sec:kernels}, and \Cref{alg:post-affine} gives the fold paths in the
same form.
W2-SEQ and W3 share the packed DRAM representation of their fold paths.
The difference is that HVX expands each weight tile to W4 before HMX
consumes it.
For $M\leq32$, one HVX thread converts the weights. For larger $M$, the
conversion is split over the HVX threads (\Cref{app:paths-vs-m}).

\Cref{tab:hvx-budget} gives the decode expansion up to seven HVX threads.
At $M=1$, expand takes $1.06$--$1.24\times$ the latency of the W3
HVX-assisted activation-fold with the same number of threads on Qwen3-0.6B
and $1.00$--$1.30\times$ on Qwen3-4B. The W2-SEQ folds and the W3 HMX-only
activation-fold use no HVX thread for weights. On Qwen3-4B, expand matches
them with three threads, except W2-SEQ at $M=32$. With seven threads it takes
$0.88$--$0.90\times$ their latency for W3 and $0.97$--$0.98\times$ for
W2-SEQ at $M\leq8$. On Qwen3-0.6B, it needs three to four threads for W3 at
$M\leq8$ and never matches the W2-SEQ folds or W3 at $M=32$. We did not
measure the energy of the multi-thread expand.

\begin{algorithm}[h]
\caption{Expand-path execution of W2-SEQ and W3 through the native W4 path.}
\label{alg:naive-w4-upcast}
\begin{algorithmic}[1]
\REQUIRE Activation codes $X^q$; packed weights in representation
  $r\in\{\text{W2-SEQ},\text{W3}\}$; per-channel $\alpha_n$,
  $b^{\mathrm{out}}_n$, and the input bias $b^{\mathrm{in}}_n$ of the represented
  weights ($\widetilde b^{\mathrm{in}}_n$ for W2-SEQ,
  $z_a\sum_kW^3_{n,k}$ for W3).
\ENSURE $Y^q\in\mathbb{Z}^{M\times N}$.
\FOR{each activation-row block and 32-channel output group $\mathcal{G}$}
  \FOR{each reduction tile along $K$}
    \IF{$r=\text{W2-SEQ}$}
      \STATE DMA: stage the $W^2$ tile; HVX: $W^4\gets2W^2+\mathbf{1}$
    \ELSE
      \STATE DMA: stage the $W^2$ and $W^1$ tiles; HVX: $W^4\gets2W^2+W^1$
    \ENDIF
    \STATE HVX: write the $W^4$ tile to VTCM in the HMX layout
    \STATE HMX: accumulate $P_{m,\mathcal{G}}$ with the native W4 product
  \ENDFOR
  \STATE HMX: $Y^q_{m,n}\gets\Requant(P_{m,n};\alpha_n,b^{\mathrm{in}}_n)$,
    $n\in\mathcal{G}$
\ENDFOR
\end{algorithmic}
\end{algorithm}

\begin{algorithm}[t]
\caption{Fold-path execution of W2-SEQ and W3 ($M\leq32$, the decode scheduler's path).}
\label{alg:post-affine}
\begin{algorithmic}[1]
\REQUIRE $X^q$; native $W^2$ (and $W^1$ for W3) tiles; $2\alpha_n$,
  $b^{\mathrm{out}}_n$, and the input biases of
  \Cref{sec:symmetric-w2,sec:hybrid-w3}.
\ENSURE $Y^q$.
\FOR{each activation-row block}
  \STATE HVX, once: $R_m$ and, if $M>1$, the correction columns carrying $L_m$ and the $R_c$ fold (W2-SEQ), or $\lfloor X^q/2\rfloor$ (W3, and at $M=1$ each chunk's W1 product $S_n$ as its tiles land).
  \FOR{each 32-channel output group, as DMA streams its tiles}
    \STATE HMX: accumulate $P^{2,q}$ (W2-SEQ, plus $L_m$ over the $S$ correction columns if $M>1$), or $\hat P^3$ with a W2
      and a W1 pass per $K$ tile (W3, \Cref{eq:w3-fused-accumulator}), at $M=1$ the W2 pass plus one tile carrying $S_n$.
    \IF{W3}
      \STATE HMX: $Y^q\gets\Requant(\hat P^3;\,2\alpha_n,\hat b^{\mathrm{in},3}_n)$.
    \ELSIF{$M=1$}
      \STATE HMX: $Y^q\gets\Requant(P^{2,q};\,2\alpha_n,(\widetilde b^{\mathrm{in}}_n-R_1)/2)$.
    \ELSE
      \STATE HMX: $Y^q\gets\Requant(P^{2,q}+L_m;\,2\alpha_n,(\widetilde b^{\mathrm{in}}_n-R_c)/2)$.
    \ENDIF
  \ENDFOR
\ENDFOR
\end{algorithmic}
\end{algorithm}

\subsection{W2-SEQ Accumulator-Fold for $2\leq M\leq32$}
\label{app:augk}

The output-correction form (\Cref{app:output-correction-form}) costs an
HVX pass over all $M\times N$ outputs and a headroom in which large
corrections are clipped. The bias-fold avoids both at $M=1$ because the single row sum
$R_1$ can enter the per-channel input bias. For $M>1$ the correction differs
per row and cannot enter a per-channel bias. The accumulator-fold
instead lets the HMX accumulation
add it through correction columns appended to both operands.

\paragraph{Algebra.}
By \Cref{eq:seq-rewrite}, the exact SEQ output is
$\Requant\bigl(P^{2}_{m,n}+R_m/2;\,2\alpha_n,\,\widetilde b^{\mathrm{in}}_n/2\bigr)$.
\Cref{eq:seq-augk} splits $R_m/2=L_m+R_c/2$ with a constant $R_c$ per call
and $L_m=(R_m-R_c)/2$. The input bias is that of the bias-fold
(\Cref{eq:seq-bias-fold}) with $R_1$ replaced by $R_c$, so the same routine
builds it, once per call for every output channel. The row term $L_m$ is
identical for every output channel, so it is the product of an activation
row whose entries sum to $L_m$ with a weight column of ones. We append $E$
tiles, i.e., $S=TE$ columns ($T=32$ on HMX), to both operands along $K$:
row $m$ of the $M\times S$ block $\Lambda$ sums
to $L_m$ and the appended weights are $+1$, a native signed W2 value. One
HMX accumulation over $K+S$ thus yields $P^{2}_{m,n}+L_m$, and one
conversion with the bias of \Cref{eq:seq-augk} gives the final code,
without an intermediate output, a headroom, or an HVX correction pass.

\paragraph{Filling the correction columns.}
Activation entries are unsigned $q$-bit codes, so a row term of up to
$(2^q-1)S$ can be spread over $S$ entries. To keep $L_m$ non-negative and
within that bound, $R_c$ is chosen per call from the row sums of the block:
with the midpoint $c=\lfloor(\min_mR_m+\max_mR_m)/2\rfloor$ and
$H=(2^{q-1}-1)S$, we set $R_c=c-2H$, so that
$L_m=\lfloor(R_m-c)/2\rfloor+H$ lies in $[0,(2^q-1)S]$ whenever the spread
$\max_mR_m-\min_mR_m$ is at most $4(2^{q-1}-1)S-4$. Entry $j$ of row $m$
holds $\lfloor L_m/S\rfloor+[\,j<L_m\bmod S\,]$, so the entries of a row
sum to $L_m$ exactly, each one is a valid code, and each lies within one of
$2^{q-1}-1+\lfloor(R_m-c)/(2S)\rfloor$, the middle of the code range plus
the row's half-deviation per entry. $E$ is the smallest power of two up to
8 that covers the spread of the block. On synthetic activations with calibration-like statistics, $E=1$
in all 4{,}500 sampled blocks ($K$ from 1024 to 12288, $M$ from 2 to 32).
The floor of an odd $R_m-c$ is the only approximation, at most half an
accumulator unit before rounding. Against the exact integer result the
measured maximum error is one 16-bit output code (two in mixed kernels with
W3 groups), versus two to nine for the output-correction form.

\paragraph{Execution.}
Once per call, HVX computes the row sums from the packed activation tiles,
writes the $\Lambda$ tiles and the constant $+1$ weight tiles into VTCM, and
folds $R_c$ into the input bias of every output channel. HMX appends the $E$ correction-column tile pairs to the accumulation of
each output tile, so an output tile takes $K/T+E$ tile products instead of
$K/T$: for $E=1$, 3\% more HMX work at $K=1024$ and 0.8\% at $K=4096$.

\paragraph{Latency.}
The accumulator-fold is the default W2-SEQ path for $2\leq M\leq32$, for
homogeneous W2-SEQ and for the W2-SEQ groups of mixed kernels. Across the 12 projection
shapes and $M\in\{2,4,8,16,32\}$ it stays within 1\% of the native W2
kernel up to $M=8$ on Qwen3-4B and 8B (at most 23\% above it, on the
smallest Qwen3-0.6B shapes from $M=8$ on).

\subsection{Operand View of the Execution Paths}
\label{app:operand-view}

\begin{figure*}[t]
  \centering
  \includegraphics[width=\textwidth]{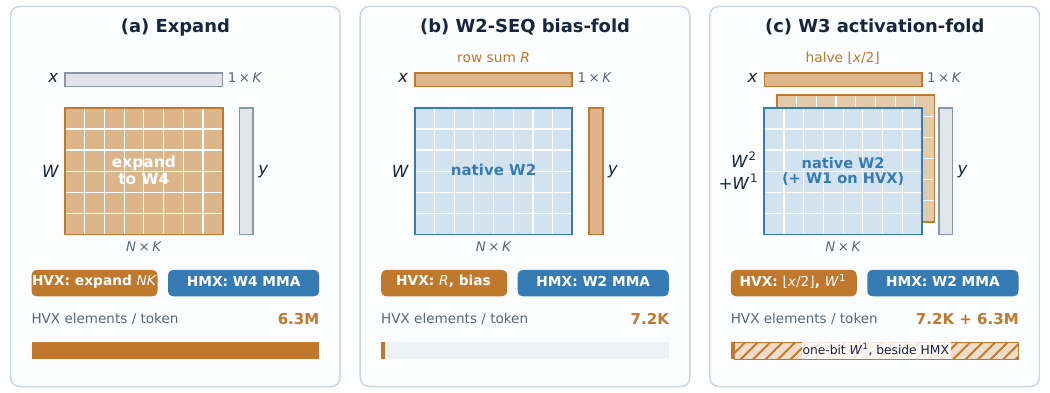}
  \caption{Operand view of the execution paths at decode size ($M=1$). Orange
  marks the data HVX must process and blue the operands HMX consumes
  natively. The expand path rewrites all $NK$ weights on HVX before HMX can start,
  whereas the W2-SEQ bias-fold touches only the activation row and the bias.
  The W3 activation-fold halves the activation row and reads the one-bit W1
  plane by table lookups beside the HMX W2 product (hatched), and its product
  joins the W2 product in one HMX accumulator that writes $y$ directly
  (\Cref{sec:hybrid-w3}). The meters give the HVX elements per token for the
  fused Up/Gate projection of Qwen3-0.6B ($K=1024$, $N=6144$) on a common
  linear scale. Operand glyphs are schematic.}
  \label{fig:work-placement-operands}
\end{figure*}

\Cref{fig:work-placement-operands} complements \Cref{fig:lowbit-schedule}
with an operand view of the paths.

\section{Measurement Methodology}
\label{app:measurement}

\subsection{Kernel Measurement Protocol}
\label{app:protocol}

Each kernel latency in \Cref{sec:eval-kernels,sec:eval-mp-latency} and \Cref{app:results}
is the median over two to five device processes, run in randomized order over
all kernels of a measurement set. Each process allocates the weights in the deployed
layout, warms up once, and averages 32 repetitions. A mixed kernel repeats a
fixed pattern of formats over its 32-channel groups (e.g., W2-SEQ, W3, W4, W4),
and its BPW is the mean of the pattern.

\subsection{End-to-End Measurement Protocol}
\label{app:e2e-protocol}

Both runtimes are measured on the same phone with the procedure of
llama.cpp's \texttt{llama-bench}. Each run is a fresh process that loads the
model, runs one warm-up pass (a prompt and one generated token), and then
five repetitions of one workload: tg128 (128 generated tokens), pp512 (a
512-token prompt), or pp512+tg128 (both). The run
value is the mean tokens per second of the five repetitions. Every
configuration and workload runs five times, and the configurations take
turns. A run starts only when the hottest CPU and NPU sensor is within
10\,$^\circ$C of its idle temperature. We report the median of the five
runs. Their spread (maximum minus minimum over the median) is at most 3.6\%
for \sysname{} and 7.5\% for llama.cpp. The llama.cpp runs use the same
procedure in separate sessions on the same phone.

\paragraph{Models.} Latencies use latency-equivalent models whose weights
are random but whose byte layout and metadata match the exported
checkpoints. The group pattern of every synthetic MP profile follows the
statistics of the real Qwen3-0.6B MP3.4 profile. The real Qwen3-4B W3 and
Qwen3-0.6B W4 checkpoints run within 0.2\% of their latency-equivalent
models at tg128, and the real Qwen3-0.6B MP3.4 checkpoint within 2\% for
all three workloads.

\paragraph{Power.} Energy is measured on a development device with the same
SoC, whose power monitors cover every rail. The sum of all rails is sampled
at 10~Hz over each run, with the screen off and no other process running,
and idle power is not subtracted. Energy is the integrated power over a run
divided by its generated tokens (end to end) or kernel calls (kernels), and
we report the median of five runs.

\subsection{llama.cpp Baselines}
\label{app:llamacpp}

\begin{table*}[t]
\centering
\caption{GGUF files of the llama.cpp baselines: quantization type of each
decoder linear projection (q, k, v, o, gate, up, down), their average BPW
including block scales, and the token-embedding / LM-head types (tied when
the model shares them). Source: the Hugging Face repositories
\texttt{unsloth/Qwen3-\{0.6B,1.7B,8B\}-GGUF},
\texttt{unsloth/Qwen3-4B-Instruct-2507-GGUF}, and
\texttt{bartowski/Qwen\_Qwen3-8B-GGUF}.}
\label{tab:gguf}
\small
\begin{tabular}{@{}lllp{0.38\linewidth}rl@{}}
\toprule
Model & Format & Source & Decoder linear tensors & BPW & Embed.\,/\,head \\
\midrule
Qwen3-0.6B & Q2\_K & unsloth & q,k,gate,up Q2\_K; v,o,down Q3\_K & 2.95 & Q6\_K (tied) \\
 & Q3\_K\_S & unsloth & all Q3\_K & 3.44 & Q6\_K (tied) \\
 & Q3\_K\_M & unsloth & q,k,gate,up Q3\_K; v Q4\_K (Q5\_K in 2 of 28 layers); o Q4\_K; down Q4\_K (Q5\_K in 1 of 28 layers) & 3.87 & Q6\_K (tied) \\
 & Q4\_0 & unsloth & q,k,v,o,gate,up Q4\_0; down Q4\_0 (Q4\_1 in 3 of 28 layers) & 4.51 & Q6\_K (tied) \\
\midrule
Qwen3-1.7B & Q2\_K & unsloth & q,k,gate,up Q2\_K; v,o,down Q3\_K & 2.93 & Q6\_K (tied) \\
 & Q3\_K\_S & unsloth & all Q3\_K & 3.44 & Q6\_K (tied) \\
 & Q3\_K\_M & unsloth & q,k,gate,up Q3\_K; v Q4\_K (Q5\_K in 2 of 28 layers); o Q4\_K; down Q4\_K (Q5\_K in 1 of 28 layers) & 3.85 & Q6\_K (tied) \\
 & Q4\_0 & unsloth & q,k,v,o,gate,up Q4\_0; down Q4\_0 (Q4\_1 in 3 of 28 layers) & 4.51 & Q6\_K (tied) \\
\midrule
Qwen3-4B-Instruct-2507 & Q2\_K & unsloth & q,k,gate,up Q2\_K; v Q4\_K; o,down Q3\_K & 2.96 & Q6\_K (tied) \\
 & Q3\_K\_S & unsloth & all Q3\_K & 3.44 & Q6\_K (tied) \\
 & Q3\_K\_M & unsloth & q,k,gate,up Q3\_K; v,down Q4\_K (Q5\_K in 2 of 36 layers); o Q4\_K & 3.85 & Q6\_K (tied) \\
 & Q4\_0 & unsloth & q,k,v,o,gate,up Q4\_0; down Q4\_0 (Q4\_1 in 4 of 36 layers) & 4.51 & Q6\_K (tied) \\
\midrule
Qwen3-8B & Q2\_K & unsloth & q,k,gate,up Q2\_K; v Q4\_K; o,down Q3\_K & 2.95 & Q2\_K\,/\,Q6\_K \\
 & Q3\_K\_S & unsloth & all Q3\_K & 3.44 & Q3\_K\,/\,Q6\_K \\
 & Q3\_K\_M & unsloth & q,k,gate,up Q3\_K; v,down Q4\_K (Q5\_K in 2 of 36 layers); o Q4\_K & 3.85 & Q3\_K\,/\,Q6\_K \\
 & Q4\_0 & bartowski & q,k,v,o,gate,up Q4\_0; down Q4\_0 (Q4\_1 in 4 of 36 layers) & 4.51 & Q4\_0\,/\,Q6\_K \\
\bottomrule
\end{tabular}

\end{table*}

The llama.cpp rows use publicly released GGUF files (\Cref{tab:gguf}). All
of them were quantized with an importance matrix computed by the publisher
on its own calibration text: Unsloth's for every file except the Qwen3-8B
Q4\_0, which Unsloth does not release and which comes from bartowski. The
throughput runs for Qwen3-4B-Instruct-2507 use the \texttt{unsloth/Qwen3-4B-GGUF}
files of the same formats, which match its GGUFs tensor by tensor in type
and size. The
k-quant presets keep some projections at higher precision: Q2\_K stores the
value, output, and down projections in Q3\_K or Q4\_K, so its decoder linear
weights average 2.93--2.96~BPW rather than the 2.63~BPW of a Q2\_K block.
We place llama.cpp at this decoder-linear BPW,
the same population of weights our BPW counts. Token embeddings and the LM
head are excluded on both sides. Our BPW counts the weight codes. The
per-group requantization parameters and the extra data of the W2-SEQ and W3 decode paths add
0.03--0.12~BPW on Qwen3-4B and 8B and up to about 0.3~BPW on 0.6B
(\Cref{tab:footprint}), whereas the GGUF BPW includes the block scales.

\subsection{Time Breakdown of the Decode Kernels}
\label{app:substage}

\begin{figure*}[t]
  \centering
  \includegraphics[width=\linewidth]{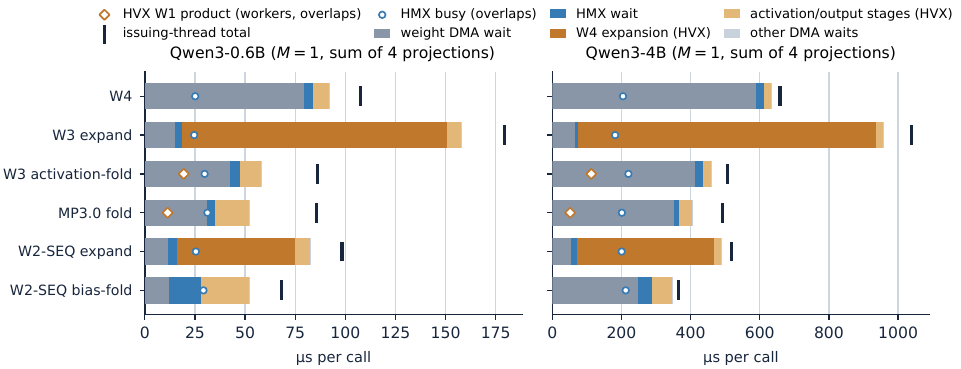}
  \caption{Time per call of the decode kernels at $M=1$ (sum of the four
  projections), measured with timing counters. Bars stack the waits and HVX
  stages of the thread that issues the DMA and HMX work. The tick marks that
  thread's total and the circle the HMX busy time, which overlaps the waits.
  The diamond marks the HVX W1 product of the fold paths, which runs on the
  worker threads and also overlaps the waits.}
  \label{fig:substage}
\end{figure*}

\Cref{fig:substage} and the timelines of \Cref{fig:lowbit-schedule} come
from a build of the same kernels with timing counters, which change the
latency by at most 1.2\%. All other latencies come from the build without
counters. \Cref{fig:substage} splits the time of the issuing thread by the
unit it waits for. In the expand paths, HVX weight conversion dominates: 132 of
179~\textmu s for W3 on Qwen3-0.6B and 861 of 1039~\textmu s on Qwen3-4B,
while HMX is busy for only 25 and 182~\textmu s. In W4 and in the
fold paths, the largest share is waiting for weight DMA, i.e.\ the
kernels are bound by weight streaming. The exception is the W2-SEQ
bias-fold on Qwen3-0.6B, whose per-call HVX stages take longer than its DMA
wait.

\section{Additional Results}
\label{app:results}

\FloatBarrier
\subsection{End-to-End Ablation of the Decode Paths}
\label{app:e2e-ablation-decode}

\begin{table}[!htbp]
\centering
\caption{End-to-end decode throughput (tg128, tokens/s) with the
fold paths (default) and with the expand path forced at decode.
Slowdown in parentheses. W4 is native and unaffected.}
\label{tab:e2e-ablation}
\resizebox{\columnwidth}{!}{%
\begin{tabular}{llrr}
\toprule
Model & Fmt & tg128 fold & expand \\
\midrule
0.6B & W4 & 142.4 & -- \\
0.6B & W3 & 157.4 & 107.2 (1.47$\times$) \\
0.6B & W2-SEQ & 168.5 & 150.7 (1.12$\times$) \\
0.6B & MP3.0 & 160.5 & 119.6 (1.34$\times$) \\
0.6B & MP3.4 & 151.2 & 124.5 (1.22$\times$) \\
\midrule
4B & W4 & 30.1 & -- \\
4B & W3 & 35.4 & 21.6 (1.64$\times$) \\
4B & W2-SEQ & 43.3 & 36.2 (1.20$\times$) \\
4B & MP3.0 & 37.6 & 28.0 (1.34$\times$) \\
4B & MP3.4 & 34.0 & 29.7 (1.14$\times$) \\
\bottomrule
\end{tabular}

}%
\end{table}

\Cref{tab:e2e-ablation} forces the \naivepath{} at decode in the end-to-end runtime.
W3 and the MP profiles then decode slower than native W4, and W2-SEQ keeps only $1.06$--$1.20\times$ the W4 throughput instead of $1.18$--$1.44\times$.

\subsection{End-to-End Throughput and Footprint}
\label{app:e2e-results}

\begin{table*}[t]
\centering
\caption{End-to-end throughput (tokens/s, median of five runs of five
repetitions each, \Cref{app:e2e-protocol}) and ratio to our W4. \sysname{} rows
use the latency-equivalent models of \Cref{app:e2e-protocol}.}
\label{tab:e1}
\small
\begin{tabular}{lllrrrrrr}
\toprule
Model & Runtime & Format & BPW & tg128 & pp512 & pp512+tg128 & tg128 / W4 & pp512 / W4 \\
\midrule
Qwen3-0.6B & \sysname{} & W2SEQ & 2.00 & 169.9 & 5686 & 653.1 & 1.19 & 0.99 \\
Qwen3-0.6B & \sysname{} & MP & 2.95 & 158.4 & 5601 & 624.1 & 1.11 & 0.97 \\
Qwen3-0.6B & \sysname{} & W3 & 3.00 & 157.8 & 5715 & 619.5 & 1.11 & 0.99 \\
Qwen3-0.6B & \sysname{} & MP & 3.44 & 151.0 & 5593 & 600.5 & 1.06 & 0.97 \\
Qwen3-0.6B & \sysname{} & MP & 3.87 & 145.6 & 5620 & 577.5 & 1.02 & 0.98 \\
Qwen3-0.6B & \sysname{} & W4 & 4.00 & 142.4 & 5753 & 580.2 & 1.00 & 1.00 \\
Qwen3-0.6B & llama.cpp & Q2\_K & 2.95 & 118.9 & 5315 & 495.2 & 0.84 & 0.92 \\
Qwen3-0.6B & llama.cpp & Q3\_K\_S & 3.44 & 122.8 & 5409 & 507.9 & 0.86 & 0.94 \\
Qwen3-0.6B & llama.cpp & Q3\_K\_M & 3.87 & 114.1 & 5398 & 475.0 & 0.80 & 0.94 \\
Qwen3-0.6B & llama.cpp & Q4\_0 & 4.51 & 115.3 & 5184 & 481.5 & 0.81 & 0.90 \\
\midrule
Qwen3-1.7B & \sysname{} & W2SEQ & 2.00 & 79.4 & 3526 & 338.3 & 1.30 & 0.99 \\
Qwen3-1.7B & \sysname{} & MP & 2.93 & 72.4 & 3438 & 313.6 & 1.19 & 0.97 \\
Qwen3-1.7B & \sysname{} & W3 & 3.00 & 68.5 & 3542 & 299.4 & 1.13 & 1.00 \\
Qwen3-1.7B & \sysname{} & MP & 3.44 & 67.3 & 3442 & 294.3 & 1.11 & 0.97 \\
Qwen3-1.7B & \sysname{} & MP & 3.85 & 63.6 & 3441 & 279.2 & 1.04 & 0.97 \\
Qwen3-1.7B & \sysname{} & W4 & 4.00 & 60.8 & 3547 & 271.6 & 1.00 & 1.00 \\
Qwen3-1.7B & llama.cpp & Q2\_K & 2.93 & 57.1 & 3023 & 253.5 & 0.94 & 0.85 \\
Qwen3-1.7B & llama.cpp & Q3\_K\_S & 3.44 & 58.1 & 3072 & 257.4 & 0.96 & 0.87 \\
Qwen3-1.7B & llama.cpp & Q3\_K\_M & 3.85 & 53.8 & 3177 & 239.5 & 0.88 & 0.90 \\
Qwen3-1.7B & llama.cpp & Q4\_0 & 4.51 & 53.1 & 3185 & 237.9 & 0.87 & 0.90 \\
\midrule
Qwen3-4B-Instruct-2507 & \sysname{} & W2SEQ & 2.00 & 43.3 & 1569 & 179.2 & 1.44 & 1.00 \\
Qwen3-4B-Instruct-2507 & \sysname{} & MP & 2.96 & 37.5 & 1543 & 160.0 & 1.25 & 0.98 \\
Qwen3-4B-Instruct-2507 & \sysname{} & W3 & 3.00 & 35.5 & 1564 & 152.1 & 1.18 & 0.99 \\
Qwen3-4B-Instruct-2507 & \sysname{} & MP & 3.44 & 34.2 & 1539 & 146.5 & 1.14 & 0.98 \\
Qwen3-4B-Instruct-2507 & \sysname{} & MP & 3.85 & 31.4 & 1542 & 136.5 & 1.05 & 0.98 \\
Qwen3-4B-Instruct-2507 & \sysname{} & W4 & 4.00 & 30.1 & 1573 & 131.5 & 1.00 & 1.00 \\
Qwen3-4B-Instruct-2507 & llama.cpp & Q2\_K & 2.96 & 27.6 & 1382 & 120.9 & 0.92 & 0.88 \\
Qwen3-4B-Instruct-2507 & llama.cpp & Q3\_K\_S & 3.44 & 28.1 & 1409 & 122.6 & 0.93 & 0.90 \\
Qwen3-4B-Instruct-2507 & llama.cpp & Q3\_K\_M & 3.85 & 25.4 & 1429 & 113.0 & 0.84 & 0.91 \\
Qwen3-4B-Instruct-2507 & llama.cpp & Q4\_0 & 4.51 & 25.3 & 1470 & 112.0 & 0.84 & 0.93 \\
\midrule
Qwen3-8B & \sysname{} & W2SEQ & 2.00 & 26.8 & 1040 & 114.8 & 1.60 & 1.00 \\
Qwen3-8B & \sysname{} & MP & 2.95 & 22.2 & 1021 & 97.2 & 1.33 & 0.98 \\
Qwen3-8B & \sysname{} & W3 & 3.00 & 21.0 & 1031 & 92.7 & 1.25 & 0.99 \\
Qwen3-8B & \sysname{} & MP & 3.44 & 19.7 & 1023 & 88.0 & 1.18 & 0.99 \\
Qwen3-8B & \sysname{} & MP & 3.85 & 17.4 & 1009 & 78.3 & 1.04 & 0.97 \\
Qwen3-8B & \sysname{} & W4 & 4.00 & 16.7 & 1038 & 75.6 & 1.00 & 1.00 \\
Qwen3-8B & llama.cpp & Q2\_K & 2.95 & 15.7 & 769.0 & 70.3 & 0.94 & 0.74 \\
Qwen3-8B & llama.cpp & Q3\_K\_S & 3.44 & 16.0 & 772.6 & 71.1 & 0.96 & 0.74 \\
Qwen3-8B & llama.cpp & Q3\_K\_M & 3.85 & 14.8 & 792.5 & 66.0 & 0.88 & 0.76 \\
Qwen3-8B & llama.cpp & Q4\_0 & 4.51 & 14.4 & 839.8 & 64.7 & 0.86 & 0.81 \\
\bottomrule
\end{tabular}

\end{table*}

\begin{table}[!htbp]
\centering
\caption{Decoder section size (GiB) of the exported models.
$^\dagger$Runs in two DSP processes.}
\label{tab:footprint}
\resizebox{\columnwidth}{!}{%
\begin{tabular}{lrrrrrr}
\toprule
Model & W2-SEQ & W3 & MP3.0 & MP3.4 & MP3.8 & W4 \\
\midrule
0.6B & 0.12 & 0.17 & 0.17 & 0.19 & 0.21 & 0.21 \\
4B & 0.89 & 1.32 & 1.33 & 1.50 & 1.66 & 1.71 \\
8B & 1.67 & 2.49 & 2.50 & 2.82 & 3.14$^\dagger$ & 3.26$^\dagger$ \\
\bottomrule
\end{tabular}

}%
\end{table}

\Cref{tab:e1} lists the end-to-end throughput behind
\Cref{fig:eval-pareto}, including prefill.

\paragraph{Weight footprint.}
The stored weights of every format are byte-proportional by construction
($\mathrm{BPW}\cdot KN/8$ per projection plus per-channel scales and
biases), so the decoder sections scale with BPW (\Cref{tab:footprint}).
A DSP process holds the decoder weights, the KV cache, and the working
buffers in one 32-bit address space, so the Qwen3-8B W4 and MP3.8 decoders
(3.1~GiB and more) run in two DSP processes and every other configuration
in one.

\FloatBarrier
\subsection{Per-Projection Decode Kernels}
\label{app:per-projection}

\begin{table}[!htbp]
\centering
\caption{W2-SEQ projection latency at decode size ($M=1$) in \textmu s.
The bias-fold is the default path.}
\label{tab:w2-m1}
\resizebox{\columnwidth}{!}{%
\begin{tabular}{llrrrrr}
\toprule
Model & Projection & W2 & Expand & Output-correction & Bias-fold & W4 \\
\midrule
Qwen3-0.6B & QKV & 18.8 & 27.7 & 20.0 & 20.8 & 30.3 \\
Qwen3-0.6B & O & 10.2 & 15.9 & 11.4 & 11.1 & 16.0 \\
Qwen3-0.6B & Up/Gate & 26.4 & 39.1 & 28.1 & 29.1 & 45.0 \\
Qwen3-0.6B & Down & 13.4 & 21.7 & 14.7 & 13.8 & 22.6 \\
 & \emph{sum} & 68.8 & 104.3 & 74.2 & 74.7 & 113.9 \\
\midrule
Qwen3-1.7B & QKV & 32.8 & 49.3 & 34.1 & 33.0 & 58.1 \\
Qwen3-1.7B & O & 17.3 & 27.4 & 18.4 & 17.5 & 30.5 \\
Qwen3-1.7B & Up/Gate & 97.0 & 126.3 & 99.9 & 97.4 & 170.9 \\
Qwen3-1.7B & Down & 46.0 & 72.9 & 48.3 & 46.2 & 85.1 \\
 & \emph{sum} & 193.2 & 275.9 & 200.7 & 194.1 & 344.6 \\
\midrule
Qwen3-4B & QKV & 59.7 & 86.5 & 61.7 & 60.0 & 109.3 \\
Qwen3-4B & O & 39.2 & 61.2 & 41.1 & 39.4 & 71.6 \\
Qwen3-4B & Up/Gate & 181.4 & 243.8 & 183.5 & 180.5 & 325.7 \\
Qwen3-4B & Down & 91.4 & 134.2 & 93.7 & 91.7 & 160.5 \\
 & \emph{sum} & 371.8 & 525.6 & 379.9 & 371.6 & 667.1 \\
\midrule
Qwen3-8B & QKV & 94.2 & 131.3 & 96.0 & 94.5 & 169.7 \\
Qwen3-8B & O & 62.5 & 94.5 & 64.3 & 62.7 & 115.2 \\
Qwen3-8B & Up/Gate & 338.2 & 488.5 & 340.2 & 339.7 & 641.0 \\
Qwen3-8B & Down & 173.5 & 261.7 & 175.4 & 173.7 & 320.9 \\
 & \emph{sum} & 668.4 & 975.9 & 675.8 & 670.6 & 1246.8 \\
\bottomrule
\end{tabular}

}%
\end{table}

\begin{table}[!htbp]
\centering
\caption{W3 projection latency at decode size ($M=1$) in \textmu s. The activation-fold (act.-fold) is
the default path. Byte-proportional act.-fold/W4 would be 0.75.}
\label{tab:w3-m1}
\resizebox{\columnwidth}{!}{%
\begin{tabular}{llrrrrr}
\toprule
Model & Projection & Expand & Act.-fold & W4 & Act.-fold/W4 & Act.-fold/Expand \\
\midrule
Qwen3-0.6B & QKV & 49.1 & 25.2 & 30.3 & 0.83 & 0.51 \\
Qwen3-0.6B & O & 26.8 & 13.7 & 16.0 & 0.86 & 0.51 \\
Qwen3-0.6B & Up/Gate & 71.7 & 36.7 & 45.0 & 0.82 & 0.51 \\
Qwen3-0.6B & Down & 37.7 & 18.1 & 22.6 & 0.80 & 0.48 \\
 & \emph{sum} & 185.3 & 93.8 & 113.9 & 0.82 & 0.51 \\
\midrule
Qwen3-1.7B & QKV & 92.3 & 45.8 & 58.1 & 0.79 & 0.50 \\
Qwen3-1.7B & O & 49.0 & 23.9 & 30.5 & 0.78 & 0.49 \\
Qwen3-1.7B & Up/Gate & 253.9 & 134.0 & 170.9 & 0.78 & 0.53 \\
Qwen3-1.7B & Down & 137.3 & 66.1 & 85.1 & 0.78 & 0.48 \\
 & \emph{sum} & 532.5 & 269.9 & 344.6 & 0.78 & 0.51 \\
\midrule
Qwen3-4B & QKV & 166.8 & 85.4 & 109.3 & 0.78 & 0.51 \\
Qwen3-4B & O & 115.8 & 55.6 & 71.6 & 0.78 & 0.48 \\
Qwen3-4B & Up/Gate & 497.9 & 249.3 & 325.7 & 0.77 & 0.50 \\
Qwen3-4B & Down & 265.1 & 125.5 & 160.5 & 0.78 & 0.47 \\
 & \emph{sum} & 1045.5 & 515.8 & 667.1 & 0.77 & 0.49 \\
\midrule
Qwen3-8B & QKV & 262.0 & 132.0 & 169.7 & 0.78 & 0.50 \\
Qwen3-8B & O & 178.6 & 89.6 & 115.2 & 0.78 & 0.50 \\
Qwen3-8B & Up/Gate & 1017.4 & 489.4 & 641.0 & 0.76 & 0.48 \\
Qwen3-8B & Down & 536.7 & 243.1 & 320.9 & 0.76 & 0.45 \\
 & \emph{sum} & 1994.7 & 954.1 & 1246.8 & 0.77 & 0.48 \\
\bottomrule
\end{tabular}

}%
\end{table}

\Cref{tab:w2-m1,tab:w3-m1} give the per-projection latencies, and
\Cref{tab:m1-agg} gives the aggregate against the DMA ceiling. Native W2
stays 7--21\% above the byte-proportional ideal (half the W4 latency), a
fixed per-kernel cost (DMA start-up, pipeline fill, per-chunk output stage)
that weighs most on the small Qwen3-0.6B shapes. The W3 activation-fold,
whose W1 product runs on HVX beside the W2 product, stays within 2--10\%
of its ideal (0.75 of the W4 latency).

\subsection{Latency versus $M$ and Path Selection}
\label{app:paths-vs-m}

\begin{figure*}[t]
  \centering
  \includegraphics[width=\textwidth]{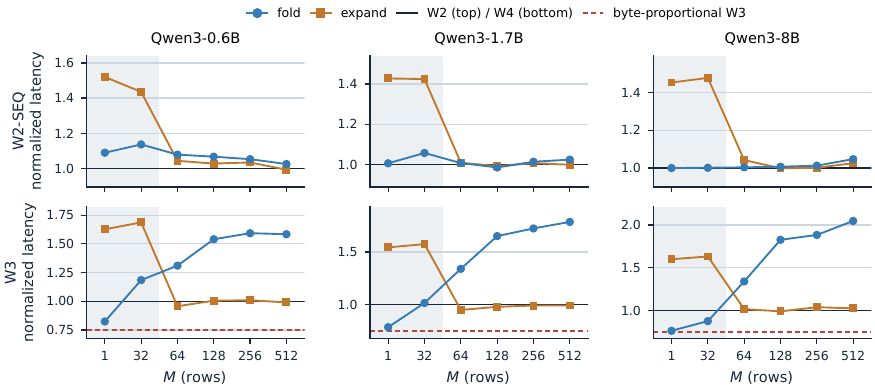}
  \caption{\Cref{fig:paths-vs-m} for Qwen3-0.6B, 1.7B, and 8B: W2-SEQ normalized to W2 (top) and W3 normalized to W4
  (bottom).}
  \label{fig:paths-vs-m-small}
\end{figure*}

\begin{table*}[t]
\centering
\caption{Aggregate (QKV+O+Up/Gate+Down) latency in \textmu s versus $M$
(values of \Cref{fig:paths-vs-m}). The runtime runs the fold paths for
$M\leq32$ and the expand path from $M=64$. The fold values at $M\geq64$
are forced. W2-SEQ/W2 uses the default path.}
\label{tab:paths-vs-m}
{\footnotesize\setlength{\tabcolsep}{3pt}%
\begin{tabular}{lrrrrrrrr}
\toprule
Model & $M$ & W2 & W2-SEQ fold & W2-SEQ expand & W2-SEQ/W2 & W4 & W3 fold & W3 expand \\
\midrule
Qwen3-0.6B & 1 & 68.6 & 74.8 & 104.3 & 1.09 & 113.9 & 93.8 & 185.3 \\
Qwen3-0.6B & 32 & 103.1 & 117.2 & 148.0 & 1.14 & 137.1 & 162.3 & 231.3 \\
Qwen3-0.6B & 64 & 161.9 & 174.7 & 168.9 & 1.04 & 196.8 & 257.8 & 188.1 \\
Qwen3-0.6B & 128 & 261.0 & 278.4 & 268.2 & 1.03 & 275.7 & 424.6 & 277.1 \\
Qwen3-0.6B & 256 & 470.5 & 495.2 & 486.5 & 1.03 & 486.2 & 774.6 & 490.3 \\
Qwen3-0.6B & 512 & 962.2 & 986.2 & 955.1 & 0.99 & 977.0 & 1547.8 & 967.0 \\
\midrule
Qwen3-1.7B & 1 & 193.2 & 194.1 & 275.9 & 1.00 & 344.6 & 269.9 & 532.5 \\
Qwen3-1.7B & 32 & 237.5 & 250.9 & 338.5 & 1.06 & 380.0 & 385.4 & 598.9 \\
Qwen3-1.7B & 64 & 381.6 & 383.9 & 384.5 & 1.01 & 470.4 & 630.3 & 445.8 \\
Qwen3-1.7B & 128 & 658.8 & 648.4 & 652.8 & 0.99 & 684.5 & 1131.2 & 669.3 \\
Qwen3-1.7B & 256 & 1216.9 & 1232.1 & 1223.9 & 1.01 & 1256.3 & 2168.2 & 1243.4 \\
Qwen3-1.7B & 512 & 2387.0 & 2442.5 & 2381.1 & 1.00 & 2406.5 & 4303.6 & 2383.0 \\
\midrule
Qwen3-4B & 1 & 372.0 & 371.7 & 525.6 & 1.00 & 667.1 & 515.8 & 1045.5 \\
Qwen3-4B & 32 & 427.4 & 440.6 & 623.6 & 1.03 & 721.3 & 695.0 & 1172.0 \\
Qwen3-4B & 64 & 684.3 & 670.2 & 685.3 & 1.00 & 860.0 & 1172.4 & 844.7 \\
Qwen3-4B & 128 & 1204.5 & 1190.7 & 1187.5 & 0.99 & 1234.9 & 2141.5 & 1202.0 \\
Qwen3-4B & 256 & 2303.5 & 2304.3 & 2269.5 & 0.99 & 2309.5 & 4277.3 & 2284.0 \\
Qwen3-4B & 512 & 4459.5 & 4568.1 & 4661.3 & 1.05 & 4680.1 & 8388.8 & 4698.3 \\
\midrule
Qwen3-8B & 1 & 672.1 & 671.8 & 975.9 & 1.00 & 1246.8 & 954.1 & 1994.7 \\
Qwen3-8B & 32 & 733.8 & 734.6 & 1083.5 & 1.00 & 1315.3 & 1155.1 & 2146.9 \\
Qwen3-8B & 64 & 1156.4 & 1160.2 & 1204.6 & 1.04 & 1557.5 & 2088.6 & 1586.1 \\
Qwen3-8B & 128 & 2106.4 & 2120.4 & 2102.8 & 1.00 & 2149.6 & 3927.2 & 2132.6 \\
Qwen3-8B & 256 & 4064.3 & 4114.1 & 4068.6 & 1.00 & 4096.2 & 7714.6 & 4261.9 \\
Qwen3-8B & 512 & 7972.2 & 8346.5 & 8176.3 & 1.03 & 8144.3 & 16669.4 & 8362.2 \\
\bottomrule
\end{tabular}
}
\end{table*}

\Cref{fig:paths-vs-m} follows both formats from decode to prefill sizes, and
\Cref{tab:paths-vs-m} lists the plotted values.
The runtime runs the fold paths for $M\leq32$. For larger $M$ it runs the
expand path, whose conversion cost
is amortized over the activation rows. At $M=32$ the W3 activation-fold runs at $0.88$ (8B) to $1.18\times$
(0.6B) W4. At $M=32$ the W2-SEQ accumulator-fold runs at $1.00$--$1.14\times$ native W2
(\Cref{app:augk}). From $M=64$ on, each linear layer is split into at least four chunks
along $N$, so that the weight DMA of one chunk overlaps HMX on the previous
one. The eight HVX threads of this NPU expand each chunk to W4, each a
contiguous range of its tiles, while HMX works on the previous chunk. The expanded W3 kernel then runs
within $5\%$ of W4 ($0.95$--$1.04\times$ on the four models for $M=64$ to
$512$), the expanded W2-SEQ kernel at $0.99$--$1.05\times$ native W2, and the
three-kind mixed kernel ($\{$W2-SEQ, W3, W4$\}$ in equal shares) at
$1.03$--$1.07\times$ W4 on Qwen3-4B and $1.10$--$1.21\times$ on 0.6B.

\paragraph{Fold paths beyond $M=32$.}
The fold points of \Cref{fig:paths-vs-m} at $M\geq64$ are forced. The runtime
does not run the fold paths beyond $M=32$. Over $M=64$ to
512 the W3 activation-fold takes $1.32$--$1.99\times$ the expand latency on
the four models. Its second HMX pass doubles the HMX work, which no longer
hides behind the weight stream once that stream is amortized over many
rows. The W2-SEQ accumulator-fold takes $0.96$--$1.04\times$ the expand
latency.

\subsection{HVX-Assisted versus HMX-Only W3 Activation-Fold}
\label{app:hvx-assisted-vs-hmx-only}

\begin{table}[!htbp]
\centering
\caption{W3 activation-fold latency versus $M$ (\textmu s, sum over QKV, O,
Up/Gate, and Down). The HVX-assisted form at $M\geq2$ is measured for
comparison only. Bold marks the fastest W3 path, and W4 is the native
reference.}
\label{tab:afold-hvx-hmx-vs-m}
\resizebox{\columnwidth}{!}{%
\begin{tabular}{lrrrr}
\toprule
Model & $M$ & HVX-assisted & HMX-only & W4 \\
\midrule
0.6B & 1 & \textbf{93.7} & 107.2 & 113.6 \\
 & 2 & 123.7 & \textbf{113.0} & 116.7 \\
 & 4 & 218.8 & \textbf{114.5} & 117.8 \\
 & 8 & 411.5 & \textbf{120.3} & 120.7 \\
 & 16 & 790.6 & \textbf{132.4} & 126.0 \\
 & 32 & 1517 & \textbf{159.8} & 137.4 \\
\midrule
1.7B & 1 & \textbf{270.3} & 308.8 & 344.0 \\
 & 2 & 324.4 & \textbf{316.3} & 349.5 \\
 & 4 & 589.2 & \textbf{319.1} & 352.3 \\
 & 8 & 1132 & \textbf{322.2} & 356.1 \\
 & 16 & 2212 & \textbf{338.8} & 363.5 \\
 & 32 & 4318 & \textbf{385.7} & 380.2 \\
\midrule
4B & 1 & \textbf{515.6} & 576.3 & 667.6 \\
 & 2 & 604.3 & \textbf{586.7} & 675.2 \\
 & 4 & 1119 & \textbf{589.1} & 679.7 \\
 & 8 & 2173 & \textbf{592.6} & 684.4 \\
 & 16 & 4257 & \textbf{614.3} & 697.1 \\
 & 32 & 8332 & \textbf{682.7} & 719.1 \\
\midrule
8B & 1 & \textbf{953.2} & 1020 & 1250 \\
 & 2 & 1058 & \textbf{1034} & 1258 \\
 & 4 & 2013 & \textbf{1036} & 1262 \\
 & 8 & 3940 & \textbf{1043} & 1269 \\
 & 16 & 7777 & \textbf{1074} & 1283 \\
 & 32 & 18545 & \textbf{1154} & 1312 \\
\bottomrule
\end{tabular}

}%
\end{table}

\Cref{tab:afold-hvx-hmx-vs-m} compares the two W3 activation-folds across $M$.
The HVX-assisted form is $0.87$--$0.93\times$ the HMX-only latency at $M=1$
but already $1.02$--$1.09\times$ at $M=2$. Each further row costs it another
HMX pass and another set of lookups, so its latency grows linearly with $M$.
The runtime therefore runs
the HVX-assisted activation-fold at $M=1$ and the HMX-only one for
$2\leq M\leq32$.

\FloatBarrier
\subsection{Comparison with T-MAN}
\label{app:tman}

\begin{table}[!htbp]
\centering
\caption{Latency of the reproduced T-MAN 2-bit kernel (asymmetric, group
size 64, 2.5~BPW) and of W2-SEQ on the \proposepath{} (2.0~BPW) on
Qwen3-4B, each summed over QKV, O, Up/Gate, and Down at $M$ activation
rows. T-MAN takes its faster path per projection, the LUT kernel once per
row or FP16 conversion with HMX. Ratio is T-MAN over W2-SEQ. Per byte
divides each latency by its weight bytes first (2.5~BPW for T-MAN, 2~bits
plus 16~B per output channel for W2-SEQ).}
\label{tab:tman}
\footnotesize\setlength{\tabcolsep}{3pt}
\begin{tabular}{@{}rlrrrr@{}}
\toprule
$M$ & Path & T-MAN (\textmu s) & W2-SEQ (\textmu s) & Ratio & Per byte \\
\midrule
1 & LUT & 419 & 372 & 1.13 & 0.92 \\
2 & LUT & 735 & 380 & 1.93 & 1.58 \\
4 & LUT & 1362 & 384 & 3.55 & 2.89 \\
8 & FP16 + HMX & 1777 & 390 & 4.56 & 3.72 \\
16 & FP16 + HMX & 1809 & 402 & 4.50 & 3.67 \\
32 & FP16 + HMX & 1930 & 439 & 4.40 & 3.59 \\
\bottomrule
\end{tabular}

\end{table}

T-MAN stores one bit-serial copy of the weights (asymmetric, group size 64,
2.5~BPW at 2 bits) and has two kernel paths. Its public LUT kernel
(precompute, linear, and finalize of its ExecuTorch fork) handles one
activation row. We run it unmodified on that layout, once per row for $M>1$
while each weight chunk is streamed into VTCM once. Its path for $M>1$ is
not public, so we implement it as its paper describes on the DMA pipeline
and thread pool of llama.cpp's Hexagon backend. HVX repacks each bit-serial
weight chunk to bit-parallel and converts it to FP16 with a two-level LUT,
and HMX multiplies in FP16, pipelined with the DMA. Both paths match a CPU
reference on every Qwen3 projection shape. \Cref{tab:tman} takes the faster
path per projection and $M$, which is the LUT kernel up to $M=4$ and the
FP16 path from $M=8$. We could not obtain correct end-to-end output from
T-MAN's public runtime on this NPU, so we compare kernels.

\subsection{Mixed Kernels per Projection}
\label{app:e3-per-projection}

\begin{figure*}[t]
  \centering
  \includegraphics[width=\textwidth]{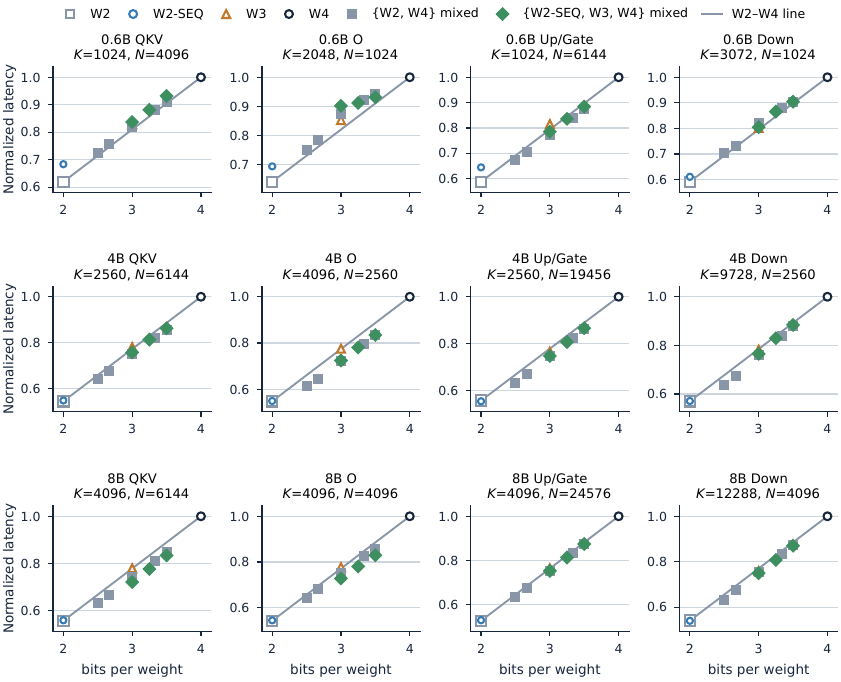}
  \caption{Mixed-precision kernel latency per projection at decode size
  ($M=1$), normalized to the W4 kernel of the same projection. Markers and line
  as in \Cref{fig:e3-interpolation}.}
  \label{fig:e3-per-projection}
\end{figure*}

\begin{table*}[t]
\centering
\caption{Deviation of mixed kernels from the composition-weighted sum of
the homogeneous kernels ($M=1$). Points are mixes $\times$ projections.
The last column is the range over the family's mixes of the aggregate
deviation.}
\label{tab:e3-summary}
{\footnotesize\setlength{\tabcolsep}{3pt}%
\begin{tabular}{llrrrrr}
\toprule
Model & Family & Mixes & Points & Mean dev. & Max $|$dev.$|$ & Agg. dev. range \\
\midrule
Qwen3-0.6B & W2–W4 & 5 & 20 & +1.0\% & 6.6\% & -0.3\,/\,+0.9\% \\
Qwen3-0.6B & W2-SEQ–W4 & 3 & 12 & -1.3\% & 7.0\% & -4.0\,/\,-1.3\% \\
Qwen3-0.6B & W3–W4 & 3 & 12 & +0.2\% & 2.6\% & -0.5\,/\,-0.1\% \\
Qwen3-0.6B & W2-SEQ/W3/W4 & 3 & 12 & +0.2\% & 6.2\% & -0.8\,/\,-0.5\% \\
\midrule
Qwen3-4B & W2–W4 & 5 & 20 & -4.4\% & 7.7\% & -5.2\,/\,-3.0\% \\
Qwen3-4B & W2-SEQ–W4 & 3 & 12 & -3.9\% & 7.5\% & -4.0\,/\,-2.6\% \\
Qwen3-4B & W3–W4 & 3 & 12 & -2.2\% & 6.3\% & -1.6\,/\,-1.1\% \\
Qwen3-4B & W2-SEQ/W3/W4 & 3 & 12 & -3.2\% & 6.5\% & -3.3\,/\,-2.4\% \\
\midrule
Qwen3-8B & W2–W4 & 5 & 20 & -2.7\% & 5.6\% & -2.7\,/\,-1.6\% \\
Qwen3-8B & W2-SEQ–W4 & 3 & 12 & -2.6\% & 5.3\% & -2.4\,/\,-1.5\% \\
Qwen3-8B & W3–W4 & 3 & 12 & -3.7\% & 7.1\% & -2.2\,/\,-1.9\% \\
Qwen3-8B & W2-SEQ/W3/W4 & 3 & 12 & -4.0\% & 7.5\% & -2.9\,/\,-2.3\% \\
\bottomrule
\end{tabular}
}
\end{table*}

\begin{table*}[t]
\centering
\caption{Aggregate latency of every mixed kernel at $M=1$ and its
composition-weighted prediction from the homogeneous kernels.}
\label{tab:e3-points}
{\footnotesize\setlength{\tabcolsep}{3pt}%
\begin{tabular}{lllrrrr}
\toprule
Model & Family & Mix & BPW & Measured (µs) & Predicted (µs) & Dev. \\
\midrule
Qwen3-0.6B & W2–W4 & w2:w2:w2:w4 & 2.50 & 80.1 & 80.1 & -0.0\% \\
Qwen3-0.6B & W2–W4 & w2:w2:w4 & 2.67 & 83.7 & 83.9 & -0.2\% \\
Qwen3-0.6B & W2–W4 & w2:w4 & 3.00 & 92.2 & 91.4 & +0.9\% \\
Qwen3-0.6B & W2–W4 & w2:w4:w4 & 3.33 & 99.0 & 98.9 & +0.1\% \\
Qwen3-0.6B & W2–W4 & w2:w4:w4:w4 & 3.50 & 102.3 & 102.6 & -0.3\% \\
Qwen3-0.6B & W2-SEQ–W4 & w2seq:w2seq:w4 & 2.67 & 84.2 & 87.7 & -4.0\% \\
Qwen3-0.6B & W2-SEQ–W4 & w2seq:w4 & 3.00 & 92.4 & 94.3 & -2.0\% \\
Qwen3-0.6B & W2-SEQ–W4 & w2seq:w4:w4 & 3.33 & 99.5 & 100.8 & -1.3\% \\
Qwen3-0.6B & W3–W4 & w3:w3:w4 & 3.33 & 99.8 & 100.3 & -0.5\% \\
Qwen3-0.6B & W3–W4 & w3:w4 & 3.50 & 103.5 & 103.7 & -0.3\% \\
Qwen3-0.6B & W3–W4 & w3:w4:w4 & 3.67 & 107.0 & 107.1 & -0.1\% \\
Qwen3-0.6B & W2-SEQ/W3/W4 & w2seq:w3:w4 & 3.00 & 93.3 & 94.0 & -0.8\% \\
Qwen3-0.6B & W2-SEQ/W3/W4 & w2seq:w3:w4:w4 & 3.25 & 98.5 & 99.0 & -0.5\% \\
Qwen3-0.6B & W2-SEQ/W3/W4 & w2seq:w3:w4:w4:w4:w4 & 3.50 & 103.4 & 104.0 & -0.6\% \\
\midrule
Qwen3-4B & W2–W4 & w2:w2:w2:w4 & 2.50 & 422.3 & 445.6 & -5.2\% \\
Qwen3-4B & W2–W4 & w2:w2:w4 & 2.67 & 446.4 & 470.2 & -5.1\% \\
Qwen3-4B & W2–W4 & w2:w4 & 3.00 & 498.4 & 519.5 & -4.1\% \\
Qwen3-4B & W2–W4 & w2:w4:w4 & 3.33 & 549.7 & 568.7 & -3.3\% \\
Qwen3-4B & W2–W4 & w2:w4:w4:w4 & 3.50 & 575.3 & 593.3 & -3.0\% \\
Qwen3-4B & W2-SEQ–W4 & w2seq:w2seq:w4 & 2.67 & 451.5 & 470.1 & -4.0\% \\
Qwen3-4B & W2-SEQ–W4 & w2seq:w4 & 3.00 & 503.0 & 519.3 & -3.2\% \\
Qwen3-4B & W2-SEQ–W4 & w2seq:w4:w4 & 3.33 & 553.8 & 568.6 & -2.6\% \\
Qwen3-4B & W3–W4 & w3:w3:w4 & 3.33 & 556.9 & 566.2 & -1.6\% \\
Qwen3-4B & W3–W4 & w3:w4 & 3.50 & 584.0 & 591.4 & -1.3\% \\
Qwen3-4B & W3–W4 & w3:w4:w4 & 3.67 & 609.8 & 616.7 & -1.1\% \\
Qwen3-4B & W2-SEQ/W3/W4 & w2seq:w3:w4 & 3.00 & 501.0 & 518.1 & -3.3\% \\
Qwen3-4B & W2-SEQ/W3/W4 & w2seq:w3:w4:w4 & 3.25 & 540.8 & 555.4 & -2.6\% \\
Qwen3-4B & W2-SEQ/W3/W4 & w2seq:w3:w4:w4:w4:w4 & 3.50 & 578.2 & 592.6 & -2.4\% \\
\midrule
Qwen3-8B & W2–W4 & w2:w2:w2:w4 & 2.50 & 791.2 & 813.0 & -2.7\% \\
Qwen3-8B & W2–W4 & w2:w2:w4 & 2.67 & 840.7 & 861.2 & -2.4\% \\
Qwen3-8B & W2–W4 & w2:w4 & 3.00 & 938.2 & 957.6 & -2.0\% \\
Qwen3-8B & W2–W4 & w2:w4:w4 & 3.33 & 1036.6 & 1054.0 & -1.7\% \\
Qwen3-8B & W2–W4 & w2:w4:w4:w4 & 3.50 & 1084.3 & 1102.2 & -1.6\% \\
Qwen3-8B & W2-SEQ–W4 & w2seq:w2seq:w4 & 2.67 & 841.5 & 862.5 & -2.4\% \\
Qwen3-8B & W2-SEQ–W4 & w2seq:w4 & 3.00 & 939.9 & 958.6 & -2.0\% \\
Qwen3-8B & W2-SEQ–W4 & w2seq:w4:w4 & 3.33 & 1038.6 & 1054.7 & -1.5\% \\
Qwen3-8B & W3–W4 & w3:w3:w4 & 3.33 & 1028.8 & 1052.3 & -2.2\% \\
Qwen3-8B & W3–W4 & w3:w4 & 3.50 & 1078.3 & 1100.9 & -2.0\% \\
Qwen3-8B & W3–W4 & w3:w4:w4 & 3.67 & 1127.7 & 1149.5 & -1.9\% \\
Qwen3-8B & W2-SEQ/W3/W4 & w2seq:w3:w4 & 3.00 & 929.5 & 957.4 & -2.9\% \\
Qwen3-8B & W2-SEQ/W3/W4 & w2seq:w3:w4:w4 & 3.25 & 1002.3 & 1029.7 & -2.7\% \\
Qwen3-8B & W2-SEQ/W3/W4 & w2seq:w3:w4:w4:w4:w4 & 3.50 & 1077.0 & 1102.1 & -2.3\% \\
\bottomrule
\end{tabular}
}
\end{table*}

\Cref{fig:e3-per-projection,tab:e3-summary,tab:e3-points} give the
measurements behind \Cref{fig:e3-interpolation} per projection and per mix,
against the composition-weighted prediction from the homogeneous kernels.

Issuing the HMX work of a mixed kernel one 32-channel group at a time,
instead of one run of same-kind groups at a time, changes its latency by at
most 0.4\% on the Qwen3-0.6B projections with $K\geq2048$.

\FloatBarrier
\subsection{Accuracy across Model Sizes}
\label{app:accuracy-size}

\begin{figure}[!htbp]
  \centering
  \includegraphics[width=\linewidth]{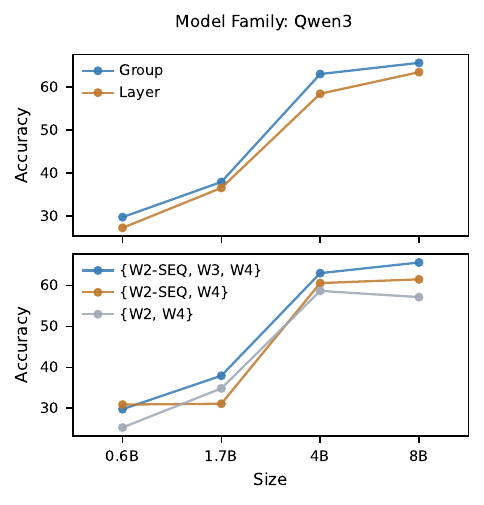}
  \caption{Accuracy (average of CSR, MMLU, and IFEval) of the weight-only
  GPTQ MP profiles at 3.0~BPW across Qwen3 model sizes. Panels as in
  \Cref{fig:prelim-mp-result}.}
  \label{fig:mp-model-size}
\end{figure}

\Cref{fig:mp-model-size} repeats the comparison of
\Cref{fig:prelim-mp-result} across the Qwen3 model sizes. The ordering
$\{$W2, W4$\}$ $<$ $\{$W2-SEQ, W4$\}$ $<$ $\{$W2-SEQ, W3, W4$\}$ holds on
Qwen3-4B and 8B. On the smaller models, where all three palettes stay at
30--38 points at 3.0~BPW, it does not: $\{$W2-SEQ, W4$\}$ falls below
$\{$W2, W4$\}$ on Qwen3-1.7B (31.1 versus 34.9), and adding W3 lowers the
average on Qwen3-0.6B (29.8 versus 30.9).

\subsection{Accuracy versus llama.cpp GGUF}
\label{app:gguf-accuracy}

\begin{table}[!htbp]
\centering
\caption{Accuracy of the llama.cpp GGUFs (rows named by their type) and of
the \sysname{} MP profile at the same BPW (MP), the pairs of
\Cref{fig:eval-pareto}. Bold marks the better value in each pair.}
\label{tab:gguf-accuracy}
\small
\begin{tabular}{@{}lrrrrrr@{}}
\toprule
Method & BPW & PPL & CSR & MMLU & IFEval & Avg. \\
\midrule
\multicolumn{7}{@{}l}{\emph{Qwen3-0.6B}} \\
Q2\_K & 2.95 & 38.60 & 40.7 & 30.8 & 40.4 & 37.3 \\
MP & 2.95 & \textbf{26.34} & \textbf{44.2} & \textbf{41.7} & \textbf{57.2} & \textbf{47.7} \\
\addlinespace[2pt]
Q3\_K\_S & 3.44 & 25.93 & 43.6 & 40.2 & 54.2 & 46.0 \\
MP & 3.44 & \textbf{23.04} & \textbf{45.9} & \textbf{44.2} & \textbf{59.1} & \textbf{49.7} \\
\addlinespace[2pt]
Q3\_K\_M & 3.87 & 22.24 & 46.3 & 43.3 & 61.4 & 50.3 \\
MP & 3.87 & \textbf{21.18} & \textbf{47.7} & \textbf{44.5} & \textbf{61.8} & \textbf{51.3} \\
\midrule
\multicolumn{7}{@{}l}{\emph{Qwen3-1.7B}} \\
Q2\_K & 2.93 & 21.37 & 49.9 & 47.9 & 55.1 & 51.0 \\
MP & 2.93 & \textbf{20.83} & \textbf{53.5} & \textbf{55.0} & \textbf{69.3} & \textbf{59.3} \\
\addlinespace[2pt]
Q3\_K\_S & 3.44 & 20.01 & 52.6 & 53.2 & 58.3 & 54.7 \\
MP & 3.44 & \textbf{18.84} & \textbf{55.3} & \textbf{57.5} & \textbf{72.9} & \textbf{61.9} \\
\addlinespace[2pt]
Q3\_K\_M & 3.85 & 16.75 & 54.4 & 56.5 & 67.9 & 59.6 \\
MP & 3.85 & \textbf{16.09} & \textbf{57.4} & \textbf{58.0} & \textbf{68.7} & \textbf{61.4} \\
\midrule
\multicolumn{7}{@{}l}{\emph{Qwen3-4B-Instruct-2507}} \\
Q2\_K & 2.96 & \textbf{11.58} & 60.6 & 65.2 & 82.6 & 69.5 \\
MP & 2.96 & 12.52 & \textbf{62.3} & \textbf{67.9} & \textbf{83.5} & \textbf{71.2} \\
\addlinespace[2pt]
Q3\_K\_S & 3.44 & \textbf{10.40} & 63.8 & 68.9 & 83.8 & 72.2 \\
MP & 3.44 & 11.21 & \textbf{64.5} & \textbf{70.0} & \textbf{84.9} & \textbf{73.1} \\
\addlinespace[2pt]
Q3\_K\_M & 3.85 & \textbf{9.99} & 65.0 & 70.4 & \textbf{85.0} & 73.4 \\
MP & 3.85 & 10.30 & \textbf{65.5} & \textbf{70.5} & 84.4 & 73.4 \\
\midrule
\multicolumn{7}{@{}l}{\emph{Qwen3-8B}} \\
Q2\_K & 2.95 & \textbf{9.99} & 63.4 & 69.8 & \textbf{85.1} & \textbf{72.8} \\
MP & 2.95 & 11.15 & \textbf{64.4} & \textbf{69.9} & 83.8 & 72.7 \\
\addlinespace[2pt]
Q3\_K\_S & 3.44 & \textbf{9.73} & 64.7 & 71.8 & \textbf{85.7} & \textbf{74.1} \\
MP & 3.44 & 10.09 & \textbf{65.7} & \textbf{72.1} & 84.1 & 74.0 \\
\addlinespace[2pt]
Q3\_K\_M & 3.85 & \textbf{9.39} & \textbf{65.6} & \textbf{73.4} & 85.1 & \textbf{74.7} \\
MP & 3.85 & 10.17 & 65.2 & 72.1 & \textbf{85.5} & 74.3 \\
\bottomrule
\end{tabular}

\end{table}

\Cref{tab:gguf-accuracy} lists the pairs of the accuracy panel of
\Cref{fig:eval-pareto}: each llama.cpp GGUF of \Cref{app:llamacpp} next to
the \sysname{} MP profile at its BPW (FlexRound weights with PrefixQuant, in
the AIMET simulation of \Cref{sec:eval-setup}). The metrics follow
\Cref{app:eval-details}. The GGUFs are also calibrated with an
importance matrix, keep some projections at a higher type (Q2\_K stores v,
o, and down in Q3\_K or Q4\_K), and quantize the embedding and the LM head, which the
decoder-linear BPW does not count.

At the BPW of Q2\_K, Q3\_K\_S, and Q3\_K\_M, the MP profiles reach a lower
perplexity and a higher Avg.\ on Qwen3-0.6B and 1.7B, by 1.0 to 10.4 points
of Avg. On Qwen3-4B-Instruct-2507 their Avg.\ is equal or higher (by up to 1.7 points) at a
higher perplexity. On Qwen3-8B the GGUFs reach the lower perplexity at every
BPW, and the two are within 0.4 points of Avg.

\subsection{Numerical Fidelity}
\label{app:fidelity}

\begin{table}[!htbp]
\centering
\caption{WikiText-2 perplexity over one 4096-token window of the real Qwen3-0.6B MP3.4-A16
checkpoint with its W2-SEQ and W3 groups on the fold paths or forced onto
the expand path. The first row scores the window one token per call and the
second 32 tokens per call, the call sizes at which the runtime selects these
fold paths.}
\label{tab:fidelity-real-ppl}
\small
\begin{tabular}{lrr}
\toprule
W2-SEQ / W3 fold path & Fold & Expand \\
\midrule
Bias-fold / HVX-assisted act.-fold & 20.675 & 20.678 \\
Accumulator-fold / HMX-only act.-fold & 20.667 & 20.660 \\
\bottomrule
\end{tabular}

\end{table}

The fold paths round slightly differently from the expand path
(\Cref{sec:hybrid-w3,app:augk}). \Cref{tab:fidelity-real-ppl} measures the
effect at model level. The runtime scores one 4096-token WikiText-2 window
of the real Qwen3-0.6B MP3.4-A16 checkpoint in calls of $M$ tokens, so every
linear runs at that $M$, with the W2-SEQ and W3 groups on the fold or the
forced expand path. The perplexity changes by less than 0.01.

\section{Accuracy Evaluation}
\label{app:limitations}

\subsection{Evaluation Setup}
\label{app:eval-details}

\begin{table}[!htbp]
\centering
\caption{Accuracy evaluation setup.}
\label{tab:eval-details}
\small
\begin{tabular}{@{}lp{0.62\columnwidth}@{}}
\toprule
Item & Setting \\
\midrule
Simulation & \raggedright AIMET GPU simulation of the A16 MP profiles (\Cref{sec:eval-setup}) \tabularnewline
Harness & \raggedright lm-evaluation-harness, thinking mode off \tabularnewline
Chat template & \raggedright IFEval only (none for WikiText-2, CSR, and MMLU) \tabularnewline
Prefix (no template) & \raggedright \texttt{<|endoftext|>\textbackslash n} (2 tokens), for WikiText-2, CSR, and MMLU \tabularnewline
Prefix (template) & \raggedright \texttt{<|im\_start|>user} (2 tokens), for IFEval \tabularnewline
WikiText-2 & \raggedright perplexity at a 4096-token context \tabularnewline
CSR & \raggedright zero-shot mean over ARC-Challenge, ARC-Easy, HellaSwag, OpenBookQA, PIQA, and WinoGrande, acc\_norm except WinoGrande (acc) \tabularnewline
MMLU & \raggedright five-shot accuracy \tabularnewline
IFEval & \raggedright zero-shot mean of its four subscores (prompt- and instruction-level, strict and loose) \tabularnewline
Avg. & \raggedright mean of CSR, MMLU, and IFEval \tabularnewline
\bottomrule
\end{tabular}
\end{table}

\Cref{tab:eval-details} lists the accuracy setup. PrefixQuant prepends a
fixed prefix whose KV cache is computed once in full precision. The
calibration, the AIMET simulation, and the on-target runs use the same
prefixes. On the NPU, the prefix KV cache (FP16, keys after RoPE) is stored
with the model and placed at the start of the KV cache, so the prefix never passes through the quantized layers
and is not scored.

\subsection{Simulated versus On-Target Perplexity}
\label{app:ontarget-ppl}

\begin{table}[!htbp]
\centering
\caption{WikiText-2 perplexity of the \sysname{} MP profiles with PrefixQuant
at the BPW of llama.cpp's Q2\_K, Q3\_K\_S, and Q3\_K\_M. On-target runs on the
NPU over the full test set (73 windows of 4096 tokens). Simulation is the AIMET GPU simulation of \Cref{sec:eval-setup}.
$\Delta$ is on-target over simulation.}
\label{tab:ontarget-ppl}
\small
\begin{tabular}{@{}lrrrr@{}}
\toprule
Model & BPW & On-target & Simulation & $\Delta$ \\
\midrule
0.6B & 2.95 & 25.88 & 26.34 & $-$1.7\% \\
0.6B & 3.44 & 22.64 & 23.04 & $-$1.7\% \\
0.6B & 3.87 & 20.94 & 21.18 & $-$1.1\% \\
\midrule
1.7B & 2.93 & 20.79 & 20.83 & $-$0.2\% \\
1.7B & 3.44 & 18.78 & 18.84 & $-$0.3\% \\
1.7B & 3.85 & 16.01 & 16.09 & $-$0.5\% \\
\midrule
4B & 2.96 & 12.57 & 12.52 & +0.4\% \\
4B & 3.44 & 11.29 & 11.21 & +0.8\% \\
4B & 3.85 & 10.36 & 10.30 & +0.6\% \\
\midrule
8B & 2.95 & 11.04 & 11.15 & $-$1.0\% \\
8B & 3.44 & 10.09 & 10.09 & +0.0\% \\
8B & 3.85 & 10.16 & 10.17 & $-$0.1\% \\
\bottomrule
\end{tabular}

\end{table}

\Cref{tab:ontarget-ppl} checks the GPU simulation behind the accuracy results against the NPU. The on-target perplexity of the \sysname{} MP profiles is within 1.8\% of the simulated one at every model and BPW.
Both sides score WikiText-2 with a 4096-token context and apply PrefixQuant (\Cref{app:eval-details}), and the on-target runs cover the full test set.
The perplexity of \Cref{app:fidelity} scores only one window of a different checkpoint and is not comparable with these values.
\par

\end{document}